\documentclass[11pt]{article}

\PassOptionsToPackage{table}{xcolor}
\usepackage[preprint]{acl}

\usepackage{times}
\usepackage{latexsym}
\usepackage[T1]{fontenc}
\usepackage[utf8]{inputenc}
\usepackage{microtype}
\usepackage{inconsolata}
\usepackage{graphicx}
\usepackage{subcaption}
\usepackage{placeins}
\usepackage{adjustbox}
\usepackage{booktabs}
\usepackage{enumitem}
\usepackage{algorithmic}
\usepackage{mathtools}
\usepackage{amsfonts}
\usepackage{nicefrac}
\usepackage{amsmath}
\usepackage{amsthm}
\usepackage{amssymb}
\usepackage{pifont}
\usepackage{array}
\usepackage{makecell}
\usepackage{multirow}
\usepackage{colortbl}
\usepackage[capitalize,noabbrev]{cleveref}

\newcounter{algorithm}
\renewcommand{\thealgorithm}{\arabic{algorithm}}

\newcommand{\algrule}{\par\noindent\hrule height 0.4pt\relax\par}
\newcommand{\algcaption}[1]{%
  \refstepcounter{algorithm}%
  \algrule
  \vskip 2pt
  \noindent\textbf{Algorithm~\thealgorithm. #1}\par
  \vskip 2pt
  \algrule
  \vskip 2pt
}
\newcommand{\algendrule}{\vskip 2pt\algrule}
\newenvironment{algbody}{%
  \par\noindent\hspace*{0.35em}%
  \begin{minipage}{\dimexpr\linewidth-0.7em\relax}%
}{%
  \end{minipage}\par
}
\newenvironment{algbox}{%
  \begin{figure}[!t]%
  \centering
  \begin{minipage}{\linewidth}%
}{%
  \end{minipage}%
  \end{figure}%
}

\usepackage{etoc}
\etocdepthtag.toc{mtchapter}
\etocsettagdepth{mtchapter}{subsection}
\etocsettagdepth{mtappendix}{none}

\newcolumntype{C}[1]{>{\centering\arraybackslash}m{#1}}
\definecolor{mildred}{RGB}{179,16,18}
\definecolor{mildblue}{RGB}{0,80,160}

\renewcommand{\arraystretch}{1.0}

\newcommand{\val}[2]{\makecell{#1\\[-2pt]\scriptsize #2}}
\newcommand{\kh}[1]{{\scriptsize #1}}
\newcommand{\attackname}{$\mathtt{M\text{-}Attack}$}
\newcommand{\ours}{$\mathtt{M\text{-}Attack\text{-}V2}$}

\theoremstyle{plain}
\newtheorem{theorem}{Theorem}

\theoremstyle{definition}

\theoremstyle{remark}

\title{Pushing the Frontier of Black-Box LVLM Attacks via\\Fine-Grained Detail Targeting}

\author{Xiaohan Zhao \quad Zhaoyi Li \quad Yaxin Luo \quad Jiacheng Cui \quad Zhiqiang Shen$^\dagger$ \\
  {\normalsize VILA Lab, Division of Computing and Mathematical Sciences, MBZUAI} \\
  {\small\texttt{\{xiaohan.zhao,zhaoyi.li,yaxin.luo,jiacheng.cui,zhiqiang.shen\}@mbzuai.ac.ae}} \\
  {\small\url{https://vila-lab.github.io/M-Attack-V2-Website/}}}

\hypersetup{
  pdftitle={Pushing the Frontier of Black-Box LVLM Attacks via Fine-Grained Detail Targeting},
  pdfauthor={Xiaohan Zhao, Zhaoyi Li, Yaxin Luo, Jiacheng Cui, Zhiqiang Shen}
}

\begin{document}

\maketitle
\begingroup
\renewcommand{\thefootnote}{}
\footnotetext{$^\dagger$Corresponding Author.}
\endgroup

\begin{abstract}
Black-box adversarial attacks on Large Vision-Language Models (LVLMs) are challenging due to missing gradients and complex multimodal boundaries. Prior state-of-the-art transfer-based approaches such as M-Attack match local crops between source and target images, but this induces high-variance, nearly orthogonal gradients across iterations. We attribute this to (i) ViT translation sensitivity, which yields spike-like gradients, and (ii) a structural asymmetry between source and target crops. We reformulate local matching as an asymmetric expectation over source transformations and target semantics, and build strong M-Attack-V2 on it. {\em Multi-Crop Alignment} (MCA) averages gradients from multiple local views per iteration, {\em Auxiliary Target Alignment} (ATA) replaces aggressive target augmentation with a small set of semantically correlated auxiliary images. {\em Patch Momentum} replays historical crop gradients, and a patch-diverse surrogate ensemble (PE+) strengthens transferable directions. M-Attack-V2 raises attack success rates on GPT-6.1 Sol from 40\% to 61\% (21\%$\uparrow$), GPT-6 Astra from 40\% to 60\% (20\%$\uparrow$), and Claude Sonnet 5.5 from 85\% to 97\% (12\%$\uparrow$).
\end{abstract}

\section{Introduction}
Large Vision-Language Models (LVLMs) have become foundational to modern AI systems, enabling multimodal tasks like image captioning~\citep{caption_2,image_caption_1,caption_3,caption_4}, VQA~\citep{q_and_a,q_and_a_2}, and visual reasoning~\citep{o3}. However, their visual modules remain vulnerable to adversarial attacks, subtle perturbations that mislead models while remaining imperceptible to humans. Prior efforts, including AttackVLM~\citep{attackvlm}, CWA~\citep{cwa}, SSA-CWA~\citep{attackbard}, AdvDiffVLM~\citep{advdiffusionvlm}, and, most effectively, the recent state-of-the-art \attackname{}~\citep{mattack}, have exploited this weakness through local-level matching and surrogate model ensembles, surpassing 90\% success rates on models like GPT-4o.

\begin{figure}[t]
    \centering
    \includegraphics[width=\linewidth]{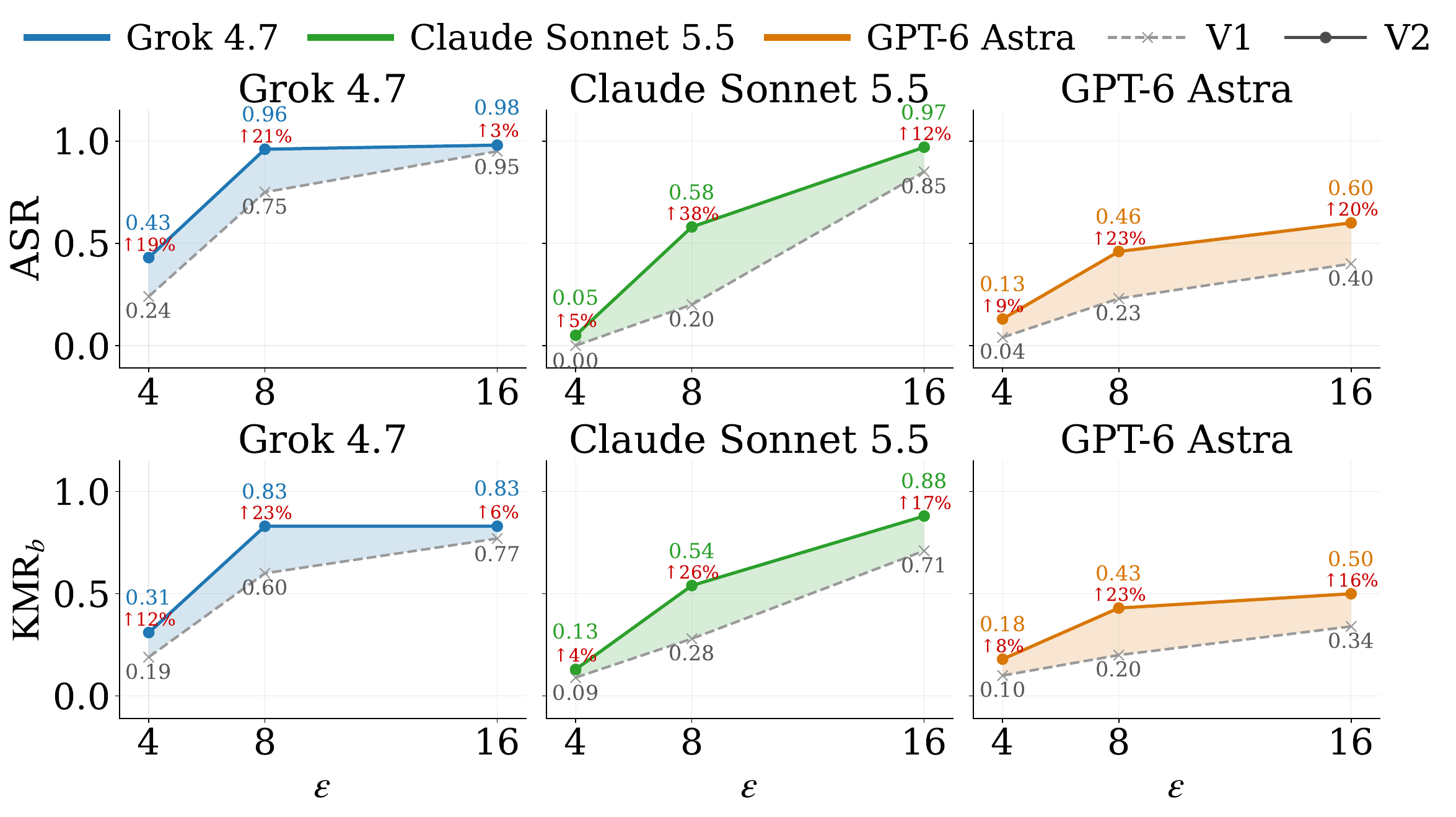}
    \caption{Improvement of \ours{} over \attackname{} on strong and up-to-date commercial black-box models (Grok 4.7, Claude Sonnet 5.5, and GPT-6 Astra). ASR and KMR stand for attack success rate and keyword matching rate, respectively.}
    \label{fig:enter-label}
\end{figure}

Despite promising performance of \attackname{}, our analysis reveals that its gradient signals are highly unstable: Even overlapping large pixel regions, two consecutive local crops share \emph{nearly orthogonal gradients}. In other words, high similarity in the pixel and embedding spaces does not translate to high similarity in the gradient space. The reason is that ViTs' gradient pattern is sensitive to translation. A tiny shift changes pixels contained in each token, altering self-attention. Moreover, patch-wise, spike-like gradient amplifies the mismatch within just a few pixels. We counter this effect by aggregating gradients from multiple crops within the same iteration, a strategy we call {\em Multi-Crop Alignment} (MCA). From a theoretical angle, MCA aggregates gradients across multiple views in a single iteration, smoothing local inconsistencies and improving cross-crop gradient stability.

We further observe that the source and target transformations in \attackname{} operate in different semantic spaces: one emphasizing extraction, the other generalization. Aggressive target augmentation introduces harmful variance. Our {\em Auxiliary Target Alignment} (ATA) mitigates this by identifying semantically similar auxiliary images to create a low-variance embedding subspace, then applying only mild shifts to enhance transferability without destabilizing the optimization.
Classic momentum is reinterpreted under this framework as {\em Patch Momentum} (PM), a replay mechanism that recycles past gradients across random crops to stabilize optimization. In parallel, we re-examine and refine \attackname{}'s model selection criterion and choose a carefully selected ensemble with diverse patch sizes to mitigate the difficulty of cross-patch transfer. We find that this selection concentrates attention more on the main object, and term it {\em Patch Ensemble$^+$} (PE$^+$) in our approach.

Together, these proposed components form the basis of our \ours{}, a robust gradient denoising framework that significantly outperforms existing black-box attack methods. Our method raises attack success rates from {\bf 40\%$\rightarrow$61\%} on GPT-6.1 Sol, {\bf 40\%$\rightarrow$60\%} on GPT-6 Astra, and {\bf 85\%$\rightarrow$97\%} on Claude Sonnet 5.5, achieving state-of-the-art performance across the board. This study not only offers a practical, modular attack strategy but also sheds light on the gradient behavior of ViT-based LVLMs under local perturbations.
To summarize, our contributions are:
\begin{itemize}
    \item We show for the first time that crop-level matching yields high-variance, near-orthogonal gradients (from ViT translation sensitivity and source/target crop asymmetry), destabilizing black-box optimization.
    \item We recast local matching as an asymmetric expectation and introduce MCA (multi-view gradient averaging) + ATA (auxiliary semantically correlated targets) to reduce variance and smooth the target manifold.
    \item We add Patch Momentum + refined PE$^+$ to amplify transferable directions, delivering large ASR gains on frontier LVLMs (e.g., GPT-6.1 Sol 40\%$\rightarrow$61\%, GPT-6 Astra 40\%$\rightarrow$60\%, Claude Sonnet 5.5 85\%$\rightarrow$97\%).
\end{itemize}

\section{Related Work}

\paragraph{Large Vision-Language Models.} Transformer-based LVLMs learn visual-semantic representations from large-scale image-text data, enabling cross-modal reasoning~\citep{corss_model_reason,corss_model_reason1,corss_model_reason2} in addition to captioning and VQA. Open-source models such as BLIP-2~\citep{blip2}, Flamingo~\citep{flamingo}, and LLaVA~\citep{llava} show strong benchmark performance. Commercial models from the GPT, Claude~\citep{claude}, Gemini~\citep{gemini}, and Grok families offer advanced reasoning and real-world adaptability, and their reasoning variants~\citep{o3} can reason over both text and images.

\paragraph{LVLM transfer-based attack.} Black-box attacks include query-based~\citep{blackbox_1,blackbox_2} and transfer-based~\citep{transfer_1,transfer_2} attacks; we focus on the latter. AttackVLM~\citep{attackvlm} pioneered transfer-based targeted LVLM attacks using CLIP~\citep{clip} and BLIP~\citep{blip} surrogates, showing that image-to-image feature matching beats cross-modal optimization, and was later adopted by subsequent works~\citep{cwa,advdiffusionvlm,attackbard,mattack}. CWA~\citep{cwa} and SSA-CWA~\citep{attackbard} extended this to Bard~\citep{gemini}: CWA improves transferability via sharpness-aware minimization~\citep{swa_1,swa_2}, while SSA-CWA adds spectrum-guided augmentation with SSA~\citep{ssa}. AnyAttack~\citep{anyattack} uses image-image matching with large-scale pretraining and fine-tuning. AdvDiffVLM~\citep{advdiffusionvlm} integrates feature matching into diffusion guidance and proposes Adaptive Ensemble Gradient Estimation (AEGE) for smoother ensemble scores. \attackname{} further surpasses these via local-level matching of source and target with an ensemble of surrogate models and diverse patch sizes, and FOA-Attack~\citep{jia2025adversarial} extends alignment from CLS to local patch tokens for additional gains. However, local-level matching still has limitations; we next introduce its background before analyzing and addressing them.

\begin{figure*}[t]
    \centering
    \begin{subfigure}{0.48\textwidth}
        \centering
        \includegraphics[width=\linewidth]{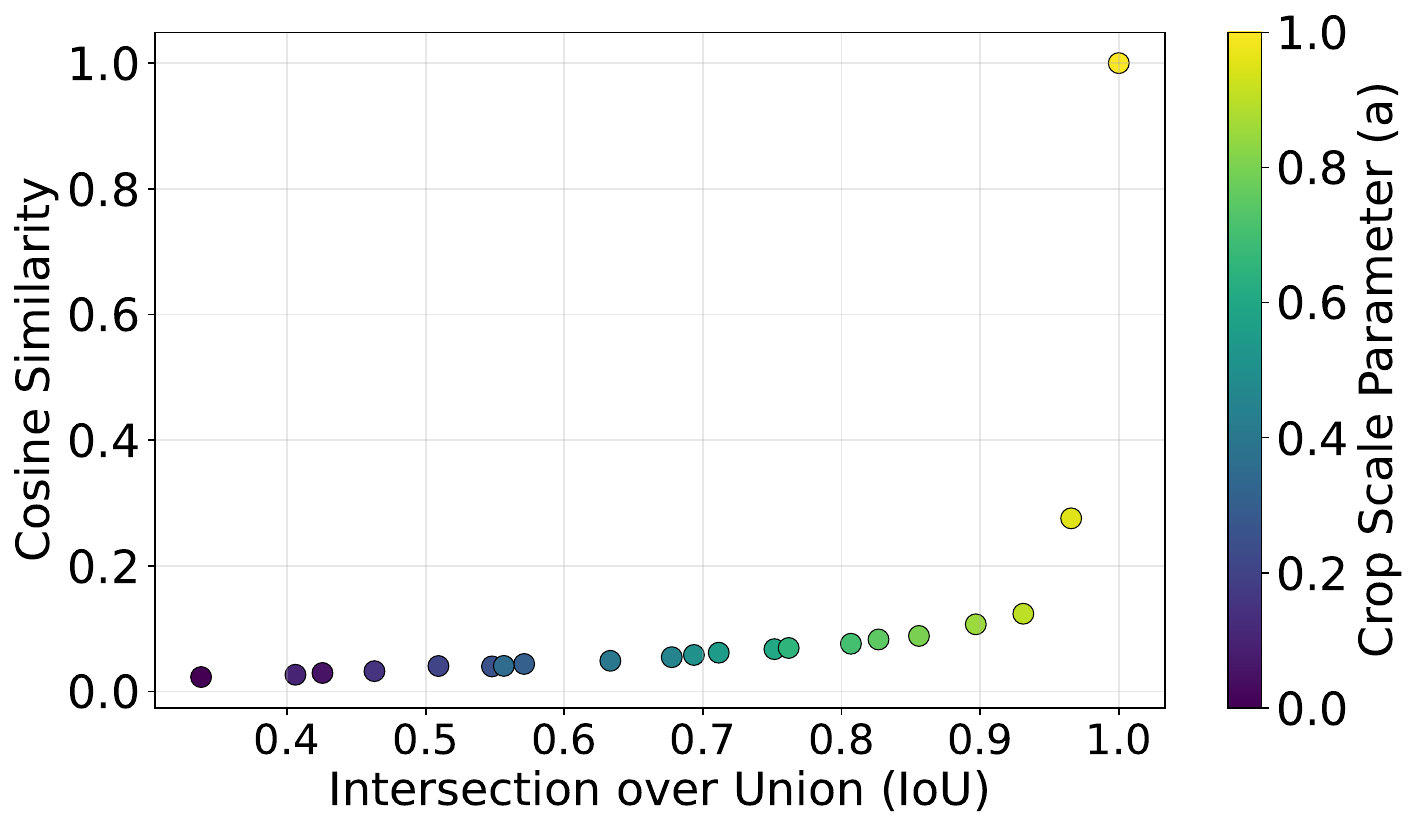}
        \caption{Source-side effect.}
        \label{fig:sim_iou}
    \end{subfigure}
    \hfill
    \begin{subfigure}{0.48\textwidth}
        \centering
        \includegraphics[width=\linewidth]{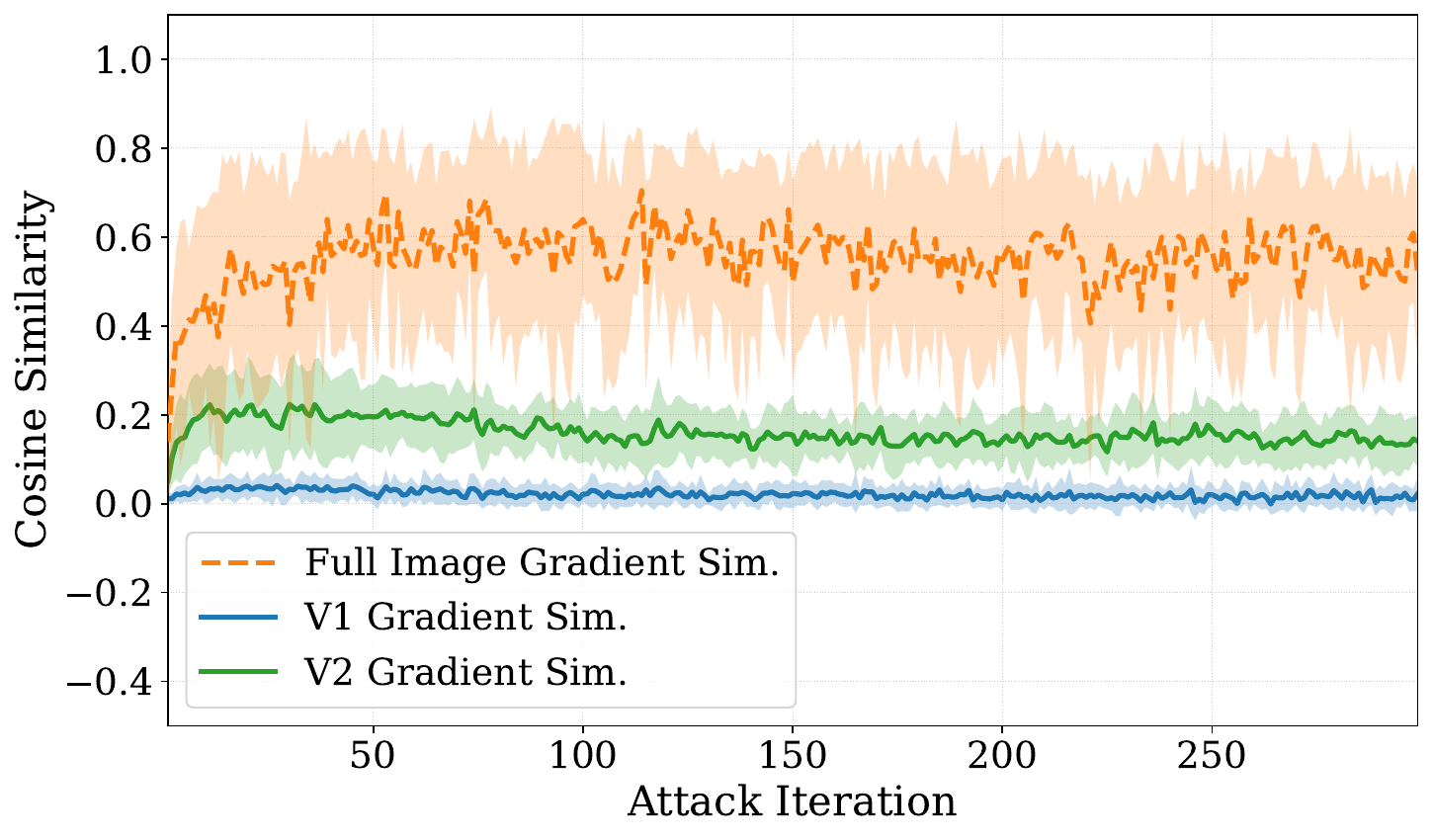}
        \caption{Full local matching.}
        \label{fig:sim_iter}
    \end{subfigure}
    \caption{Gradient mismatch in local-level matching.
(a)~Isolating the source-side effect (target fixed):
gradient similarity decays rapidly and falls below $0.1$ once IoU $< 0.8$, despite substantial pixel overlap. (b)~Combined effect in \attackname{} (both crops vary per iteration): inter-iteration gradient similarity is near zero; \ours{} improves it to around $0.2$. Results are averaged over 200 runs.}
    \label{fig:overall-comparison}
\end{figure*}

\section{Approach}
\subsection{Limitations in Local-level Matching-Based Methods}

\paragraph{Local-level matching in \attackname{}~\citep{mattack}.} Consider a clean source image $\tilde{\mathbf{X}}_{\text{sou}}$ and a target image $\mathbf{X}_{\text{tar}}$. The objective of black-box transfer attacks is to minimally perturb the source image by $\delta$ so that the perturbed image $\mathbf{X}_{\text{sou}} = \tilde{\mathbf{X}}_{\text{sou}} + \delta$ aligns semantically with the target under an inaccessible black-box model $f_{\xi}$. Due to the inaccessibility of $f_{\xi}$, surrogate models $f_{\phi}$ approximate the semantic alignment via cosine similarity:
\begin{equation} \label{equ:matching}
\begin{aligned}
&\arg\max_{\mathbf{X}_{\text{sou}}}\mathrm{CS}(f_{\phi}(\mathbf{X}_{\text{sou}}), f_{\phi}(\mathbf{X}_{\text{tar}})) \\
&\quad \text{s.t.} \quad \lVert \delta \rVert_p \le \epsilon,
\end{aligned}
\end{equation}
where $\mathrm{CS}$ denotes cosine similarity.
\attackname{} enhances Eq.~(\ref{equ:matching}) using \textit{local-level matching}. At iteration $i$, it applies predefined local transformations $\mathcal{T}_s$ and $\mathcal{T}_t$ to extract local area $ \hat{\mathbf{x}}^s_i $ from the source $\mathbf{X}_\mathrm{sou}$ and $ \hat{\mathbf{x}}^{t}_i $ from the target $\mathbf{X}_\mathrm{tar}$, respectively. These transformations satisfy essential properties, such as spatial overlap and diversified coverage of extracted local regions $\{ \hat{\bf x}_i\}$~\citep{mattack}. Formally, the local-level matching optimizes:
\begin{equation}
\mathcal{M}_{\mathcal{T}_s, \mathcal{T}_t} = \mathbb{E}_{f_{\phi_j} \sim \phi}[\text{CS}(f_{\phi_j}(\hat{\mathbf{x}}^s_i), f_{\phi_j}(\hat{\mathbf{x}}^{t}_i))],
\end{equation}
where $f_{\phi_j}$ is sampled from an ensemble of surrogate models $\phi$. Intuitively, matching local image regions instead of entire images enhances the semantic precision of perturbations by directing optimization towards semantically significant details. Despite its effectiveness, \attackname{} encounters a critical challenge of \textit{unexpectedly low} gradient similarity, which we investigate in detail next.

\paragraph{Extremely low gradient overlap.}
In \attackname{} two random crops
$\hat {\bf x}_i^s$ and $\hat {\bf x}_i^t$ are
matched at every iteration.
One would expect the gradients inside the shared region of two successive source crops, i.e.,
$\bigl(\nabla_{\hat {\bf x}_i^s} \mathcal{M}_{\mathcal{T}_s, \mathcal{T}_j},\nabla_{ \hat{\bf x}_{i+1}^s}   \mathcal{M}_{\mathcal{T}_s, \mathcal{T}_j}\bigr)$, to correlate,
because the underlying pixels partly coincide.
Surprisingly, Fig.~\ref{fig:sim_iter} shows the opposite:
their cosine similarity is \textbf{\textit{almost zero}}. We explain this unexpectedly low gradient overlap mainly with two factors:
\textit{ViTs' inherent translation sensitivity} and \textit{an overlooked asymmetry} in the local matching framework.

\begin{figure*}[t]
    \centering
    \begin{subfigure}[b]{0.36\textwidth}
        \centering
        \includegraphics[width=\linewidth,trim=20 26 22 4,clip]{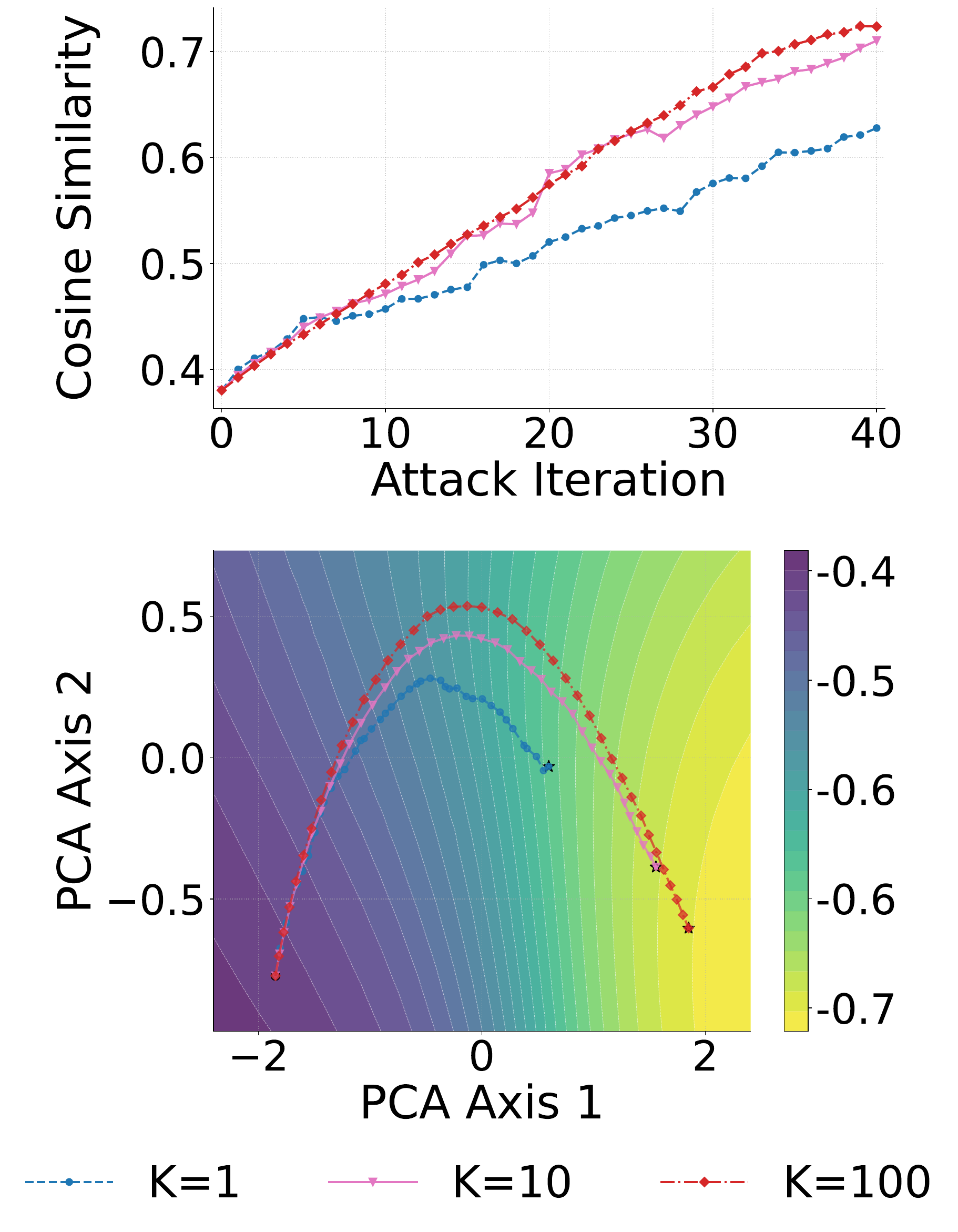}
        \caption{Optimization trajectories.}
        \label{fig:trajectory}
    \end{subfigure}
    \hspace{0.04\textwidth}
    \begin{subfigure}[b]{0.561\textwidth}
        \centering
        \includegraphics[width=\linewidth]{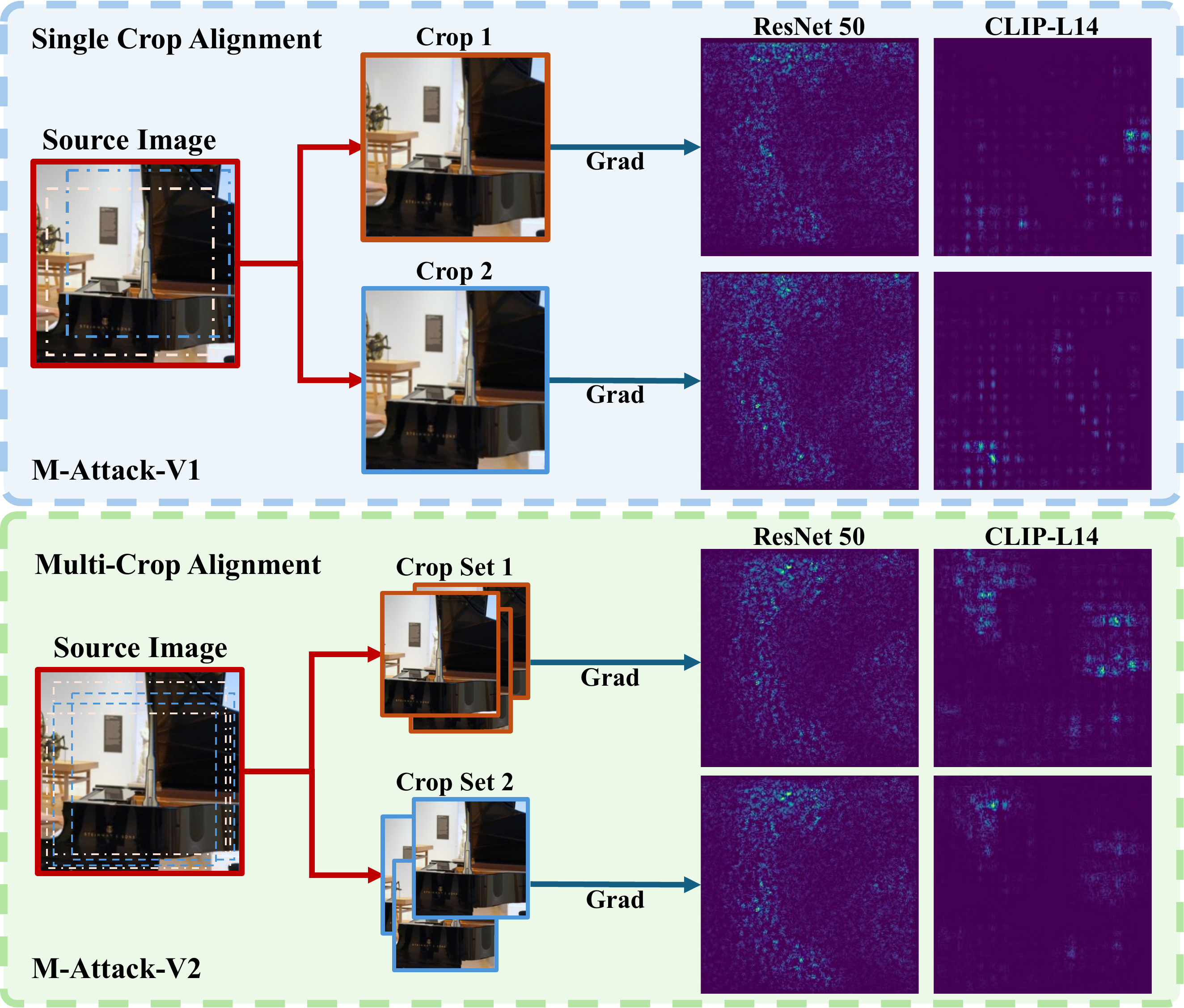}
        \caption{Gradient patterns.}
        \label{fig:gradient pattern}
    \end{subfigure}
    \caption{Comparison of: (a) optimization trajectories under different $K$; (b) gradient patterns of single-crop alignment and multi-crop alignment (MCA). The gradient pattern of ResNet-50 remains consistent when large regions overlap, while the gradient pattern of ViTs changes dramatically. MCA helps smooth this effect, improving gradient similarity and accelerating convergence.}
    \label{fig:mca comparison}
\end{figure*}

First, we keep the \textit{same target crop} within a fixed iteration, draw two random source crops, and measure their gradient similarity (Fig.~\ref{fig:sim_iou}). Our finding reveals an exponential decay that quickly plateaus below $0.1$ even with 60\% overlap. Source cropping alone breaks gradient stability. The root cause is ViTs' patch-wise, spike-like gradient pattern, which is highly sensitive to translation. ViTs tokenize images on a fixed, non-overlapping grid. Even sub-pixel shifts change each patch's token mix. These token changes ripple through self-attention, altering weights and redirecting gradients for \emph{all} tokens, so the resulting pixel-level gradient pattern diverges sharply. Worse, gradient magnitudes are uneven. Therefore, even similar patterns that miss a few pixels can break gradient similarity (Fig.~\ref{fig:gradient pattern}).

Then, the situation in \attackname{} is further aggravated by the target side. Radical target cropping in \attackname{} further destabilizes the optimization,
because source and target crops play fundamentally \textbf{\textit{asymmetric}} roles. Cropping the source acts in \textit{pixel space}:
it rearranges patch embeddings and attention weights,
producing gradients that reflect different local views of the image. Cropping the target, by contrast, acts in \textit{feature space}: it translates the reference embedding that the optimization aligns toward. The former sculpts the perturbation, while the latter moves the goalpost. \attackname{} adopts the same aggressive cropping strategy for both branches;
however, what diversifies source-side views
introduces high variance in the reference embedding on the target side, destabilizing the optimization goal. Fig.~\ref{fig:sim_iter} confirms this compounded effect: \attackname{}'s inter-iteration gradient similarity remains near zero throughout training.

\subsection{Asymmetric Matching over Expectation}

The analysis above turns local matching into two coupled estimation problems: estimating a source-side direction supported by multiple semantic-preserving local views, and selecting target-side references that cover meaningful target variations while remaining close to the intended embedding. We therefore reformulate local matching as an \emph{asymmetric expectation} over source transformations and target semantics:
\begin{equation}
    \min_{\lVert \delta \rVert_p \le \epsilon} \mathbb{E}_{\mathcal{T} \sim \mathcal{D},y \sim \mathcal{Y}} \left[ \mathcal{L}(f(\mathcal{T}({\bf X}_{\text{sou}})), y) \right], \label{equ:new_frame}
\end{equation}
where $\mathcal{D}$ denotes source-side local transformations and $\mathcal{Y}$ denotes target-side semantic references. The two expectations act in different spaces. $\mathcal{D}$ changes pixels and therefore which view of the perturbed image receives the update; $\mathcal{Y}$ changes embeddings and therefore the goal that view is aligned to. \attackname{} samples both with the same aggressive crop, one sample per iteration, so neither expectation is estimated well. MCA, ATA, and PM instantiate these terms: MCA averages source crops to recover a crop-consistent direction, ATA builds a semantic reference set with auxiliary anchors, and PM accumulates recurring local evidence across iterations (Fig.~\ref{fig:full compare}).

\subsection{Multi-Crop Alignment for Crop-Consistent Optimization} \label{sec:MCA}

We model the gradient from a local transformation $\mathcal{T}$ as
\begin{equation}
    g_{\mathcal{T}} = \mu + \xi_{\mathcal{T}},
\end{equation}
where $\mu=\mathbb{E}_{\mathcal{T}\sim\mathcal{D}}[g_{\mathcal{T}}]$ denotes the crop-consistent adversarial direction, and $\xi_{\mathcal{T}}$ denotes the crop-specific component introduced by a particular crop realization. The shared component captures the target-injection signal that persists across semantic-preserving local views, while the residual varies with token boundaries, crop positions, and attention changes.

Under this model, averaging multiple crops helps recover the semantically consistent component. When $K$ crop gradients are accumulated, the shared component adds coherently as $K\mu$. The crop-specific residuals are weakly correlated across local views, so their accumulated norm grows more slowly. After averaging, the estimator preserves $\mu$ and attenuates the residual:
\begin{equation}
  \begin{aligned}
     \hat g_K
    &=
    \nabla_{\bf X_{\text{sou}}} \hat{\mathcal{L}}({\bf X_{\text{sou}}}) \\
    & =
    \frac{1}{K} \sum_{k=1}^{K} \nabla_{\bf X_{\text{sou}}} \mathcal{L}(f(\mathcal{T}_k({ \bf X}_{\text{sou}})), y)  \\
     & =
    \mu + \frac{1}{K}\sum_{k=1}^{K}\xi_{\mathcal{T}_k}.
 \end{aligned}
  \label{equ:multiple-matching}
  \end{equation}
Therefore, Multi-Crop Alignment (MCA) estimates the adversarial direction jointly supported by multiple local views, which is also supported by the following theorem (proof in \cref{sec: theory}):

\begin{theorem} \label{thm:variance}
Let $g_k=\nabla_{\mathbf{X}_\text{sou}} \mathcal{L}(f(\mathcal{T}_k(\mathbf{X}_\text{sou})), y)$ denote the gradient from $\mathcal{T}_k$, with mean $\mu=\mathbb{E}[g_k]$ and variance $\sigma^2=\mathbb{E}[\lVert g_k-\mu \rVert_2^2]$. Let $\overline{\rho}$ denote the average pairwise correlation between crop-specific residuals $g_k-\mu$. The gradient variance from $K$ averaged crops is bounded by:
\begin{equation}
    \operatorname{Var} \bigg (\frac{1}{K}\sum_{k=1}^K g_k \bigg) \le \frac{\sigma^2}{K}  + \frac{K-1}{K}\overline{\rho}\sigma^2\label{equ:bound}.
\end{equation}
\end{theorem}
\paragraph{Insight.}
The bound separates two effects in MCA. The diagonal term $\sigma^2/K$ is the part that averaging can suppress: it corresponds to crop-specific fluctuations that do not consistently reappear across local views. The correlation term $\frac{K-1}{K}\overline{\rho}\sigma^2$ is the part that remains when crops share a common direction, since correlated components accumulate coherently instead of vanishing under averaging. Thus, increasing $K$ weakens view-specific artifacts while preserving directions repeatedly supported by different crops. Empirically, a modest number of crops ($K=10$) is sufficient to improve the update direction, with additional margin from $K=100$ (\cref{fig:trajectory}, \cref{appendix:other params}).

This crop-consistent estimate also alleviates the known translation sensitivity of ViTs. As shown in Fig.~\ref{fig:gradient pattern}, MCA keeps high-activity regions more stable than single-crop alignment. Consequently, inter-iteration gradient similarity increases from near zero in \attackname{} to around $0.2$ in \ours{} (Fig.~\ref{fig:sim_iter}).

\subsection{Reference Manifold Coverage via \textit{Auxiliary Target Alignment}} \label{subsec:ata}

Because a target crop changes the reference embedding rather than the optimized view, the radical target transformations in \attackname{} can be understood as an indirect way to sample multiple points on the target-side reference manifold. Their benefit comes from expanding the coverage of the alignment target beyond a single rigid embedding.

However, pixel-space radical transformations are a noisy sampler of this manifold.
They may produce useful reference points, but they can also remove key semantic evidence, emphasize background regions, or move the embedding toward an unrelated local view.
In such cases, the optimization target changes in the embedding space without providing meaningful semantic support.
A cleaner solution is to sample additional support points directly near the target semantics in the embedding space.

To this end, we introduce $P$ auxiliary images $\{\mathbf{X}_\text{aux}^{(p)} \}_{p=1}^P$ as an \textbf{auxiliary set} that acts as additional anchors, collectively forming a richer sub-manifold of aligned embeddings. During each update, we apply a \emph{mild} random transformation $\tilde{\mathcal{T}} \sim \tilde{\mathcal{D}}$ to every anchor, preserving reference manifold coverage while avoiding large reference drift. Let ${y}_0 = f(\hat{\mathcal{T}}_0(\mathbf{X}_\text{tar}))$,  $\tilde{y}_p = f(\tilde{\mathcal{T}}_p(\mathbf{X}_\text{aux}^{(p)}))$ denote sampled semantics in one iteration. The objective $\hat{\mathcal{L}}$ in Eq.~(\ref{equ:multiple-matching}) becomes:
\begin{equation} \begin{aligned}
&\hat{\mathcal{L}} = \frac{1}{K} \sum_{k=1}^K \Big[ \mathcal{L}(f(\mathcal{T}_k(\mathbf{X}_\text{sou})), y_0) \\
&\ \ + \frac{\lambda}{P} \sum_{p=1}^P \mathcal{L}(f(\mathcal{T}_k(\mathbf{X}_{\mathrm{sou}})), \tilde{y}_p) \Big],
\end{aligned} \end{equation}
where $\lambda\in[0,1]$ interpolates between the original target and its auxiliary neighbors. $\lambda=0$ reduces to \attackname{} local-local matching with a single target. ATA expands reference manifold coverage while keeping the sampled anchors close to the intended target semantics, providing low-variance, semantics-preserving updates. The auxiliary set can be built in various ways, e.g., via image-image retrieval or diffusion methods. We now compare the embedding drift induced by radical pixel-space transformations and auxiliary semantic anchors under three mild assumptions: (A1)~the surrogate $f$ is $L$-Lipschitz, $\lVert f(y) - f(x)\rVert \le L \Vert y - x \rVert$; (A2)~auxiliary data $\mathbf{X}_{\text{aux}}^{(p)}$ retrieved via semantic similarity to $\mathbf{X}_{\text{tar}}$ satisfy $\mathbb{E}[\lVert f(\mathbf{X}^{(p)}_{\text{aux}}) - f(\mathbf{X}_{\text{tar}})\rVert] \le \delta$ (justified in \cref{appendix:justification}); and (A3)~a random transformation $\mathcal{T} \sim D_\alpha$ has bounded pixel-level distortion, $\mathbb{E}[\lVert \mathcal{T}(\mathbf{X})-\mathbf{X} \rVert] \le \alpha$.

\begin{theorem}
\label{thm:drift}
    Let $\mathcal{T}\sim D_\alpha$ denote the transformation used in \attackname{}, and $\tilde{\mathcal{T}}\sim D_{\tilde{\alpha}}$ with $\tilde{\alpha}\ll \alpha$ the transformation in \ours{}. Define \textbf{embedding drift} of transformation $\mathcal{T}$ applied to $\mathbf{X}$ on model $f$ as: $\Delta_{\rm{drift}}(\mathcal{T};\mathbf{X}) := \mathbb{E}_{\mathcal{T}}[\lVert f(\mathcal{T}(\mathbf{X})) - f(\mathbf{X}_{\mathrm{tar}}) \rVert]$. Then, we have:
    \begin{equation}
    \begin{aligned}
    &\Delta_{\rm{drift}}(\mathcal{T};\mathbf{X}_{\mathrm{tar}}) \le L\alpha, \\ &\Delta_{\rm{drift}}(\tilde{\mathcal{T}};\mathbf{X}_{\rm{aux}}^{(p)}) \le L\tilde{\alpha} + \delta.
    \end{aligned}
\end{equation}
\end{theorem}

\paragraph{Insight.}
The bound separates two ways to expand the target-side reference manifold. Radical target transformations move the reference through pixel space, so their embedding drift is controlled by $L\alpha$; even a moderate pixel distortion can become large after passing through a sensitive surrogate encoder. ATA instead uses auxiliary anchors that are already close to the target in embedding space, contributing a semantic-neighborhood term $\delta$, and then applies only mild transformations with $\tilde{\alpha}\ll\alpha$. Thus, ATA increases reference coverage through semantically aligned support points while keeping the reference drift bounded by $L\tilde{\alpha}+\delta$ (proof in \cref{sec:proof-drift}).

\subsection{Temporal Consensus Accumulation via \textit{Patch Momentum}}

Momentum, introduced in MI-FGSM~\citep{transfer_1}, is widely adopted in transfer attacks. In our framework, its role becomes more specific: each gradient is an MCA/ATA estimate built from sampled local views and reference anchors. Patch Momentum replays these historical crop estimates across iterations, extending the consensus accumulation in Sec.~\ref{sec:MCA} from the crop dimension to the temporal dimension.

Define the momentum buffer as $m_r=\beta_1 m_{r-1}+(1-\beta_1)g_r$, where $g_r=\nabla_{\mathbf{X}_{\mathrm{adv}}}\hat{\mathcal{L}}_r$ is the MCA/ATA gradient estimate at iteration $r$. Let $g_r=\mu_r+\xi_r$, where $\mu_r$ is the local-view consensus direction near the current adversarial image and $\xi_r$ is the crop-sensitive residual. Unrolling the EMA gives
\begin{equation}
\begin{aligned}
    m_r={}&(1-\beta_1)\sum_{\tau=0}^{r}\beta_1^\tau \mu_{r-\tau} \\
    &+(1-\beta_1)\sum_{\tau=0}^{r}\beta_1^\tau \xi_{r-\tau}.
\end{aligned}
\end{equation}
Although historical gradients are partially stale after each projected update, they remain informative along a small-step adversarial trajectory. Neighboring iterations often share similar consensus directions, so the first term accumulates coherently. Crop-sensitive residuals vary with sampled crops, token boundaries, and target references; their temporal accumulation grows more slowly under EMA. Thus, Patch Momentum increases the effective crop coverage across iterations without additional backward passes.

The Adam-style~\citep{adam} second moment, $v_r=\beta_2 v_{r-1}+(1-\beta_2)g_r^2$, further normalizes spike-shaped ViT gradients and prevents a few high-magnitude coordinates from dominating the accumulated direction. We therefore term this update \textit{Patch Momentum}: in addition to the standard acceleration effect of momentum, it acts as a temporal replay mechanism for patch-level consensus.
Algorithm~\ref{alg:mattack_v2} details the MCA, ATA, and PM update loop, with red terms marking the additions over \attackname{}. Fig.~\ref{fig:full compare} in \cref{appendix:alg} visualizes the corresponding \attackname{} and \ours{} local-matching pipelines, and the final projected update uses PGD~\citep{pgd} with ADAM~\citep{adam}.

\begin{algbox}
\algcaption{\ours{}}
\label{alg:mattack_v2}
\footnotesize
\begin{algbody}
\begin{algorithmic}[1]
\REQUIRE $\mathbf{X}_{\mathrm{clean}}$, $\mathbf{X}_{\mathrm{tar}}$, \textcolor{mildred}{auxiliary set $\mathcal{A}$}, \textcolor{mildred}{patch ensemble$^+$ $\Phi^+$}, iterations $n$, step size $\alpha$, budget $\epsilon$, crops $K$, auxiliary weight $\lambda$.
\ENSURE Adversarial image $\mathbf{X}_{\mathrm{adv}}$.
\STATE $\mathbf{X}_{\mathrm{adv}}\gets\mathbf{X}_{\mathrm{clean}}$.
\FOR{$i=1$ to $n$}
    \STATE \textcolor{mildred}{Draw $K$ transforms} $\{\mathcal{T}_k\}_{k=1}^{K}\sim\mathcal{D}$; set $g\gets\mathbf{0}$.
    \FOR{$k=1$ to $K$ \textcolor{mildblue}{in parallel}}
        \STATE Draw $\{\tilde{\mathcal{T}}_p\}_{p=0}^{P}\sim\tilde{\mathcal{D}}$.
        \FOR{$j=1$ to $m$}
            \STATE $z_{k,j}\gets f_{\phi_j}(\mathcal{T}_k(\mathbf{X}_{\mathrm{adv}}))$.
            \STATE $y_0\gets f(\tilde{\mathcal{T}}_0(\mathbf{X}_{\mathrm{tar}}))$.
            \STATE \textcolor{mildred}{$y_p\gets f(\tilde{\mathcal{T}}_p(\mathbf{X}_{\mathrm{aux}}^{(p)}))$ for $p=1,\dots,P$.}
            \STATE $\hat{\mathcal{L}}_k\gets\mathcal{L}(z_{k,j},y_0)+\textcolor{mildred}{\frac{\lambda}{P}\sum_{p=1}^{P}\mathcal{L}(z_{k,j},y_p)}$.
            \STATE $g\gets g+\frac{1}{Km}\nabla_{\mathbf{X}_{\mathrm{adv}}}\hat{\mathcal{L}}_k$.
        \ENDFOR
    \ENDFOR
    \STATE Update $\mathbf{X}_{\mathrm{adv}}$ from $g$ with \textcolor{mildred}{Patch Momentum}.
\ENDFOR
\RETURN $\mathbf{X}_{\mathrm{adv}}$.
\end{algorithmic}
\end{algbody}
\algendrule
\end{algbox}

\begin{table*}[t]
\centering
\renewcommand\arraystretch{1.05}
\begin{adjustbox}{width=0.96\textwidth,center}
\small
\setlength{\tabcolsep}{1.8pt}
\begin{tabular}{c|c|cccc|cccc|cccc|cc}
\toprule
\multirow{2}{*}{\textbf{Method}} & \multirow{2}{*}{\textbf{Model}} &
\multicolumn{4}{c|}{\textbf{GPT-6 Astra}} &
\multicolumn{4}{c|}{\textbf{GPT-6.1 Sol}} &
\multicolumn{4}{c|}{\textbf{Claude Sonnet 5.5}} &
\multicolumn{2}{c}{\textbf{Imperceptibility}} \\
\cmidrule{3-16}
& & $\text{KMR}_a$ & $\text{KMR}_b$ & $\text{KMR}_c$ & ASR
  & $\text{KMR}_a$ & $\text{KMR}_b$ & $\text{KMR}_c$ & ASR
  & $\text{KMR}_a$ & $\text{KMR}_b$ & $\text{KMR}_c$ & ASR
  & $\ell_1\!\downarrow$ & $\ell_2\!\downarrow$ \\
\midrule
\multirow{3}{*}{AttackVLM~\citep{attackvlm}} & \textbf{B/16} & 0.14 & 0.03 & 0.00 & 0.01 & 0.14 & 0.05 & 0.00 & 0.01 & 0.16 & 0.08 & 0.00 & 0.02 & 0.034 & 0.040 \\
 & \textbf{B/32} & 0.14 & 0.07 & 0.00 & 0.02 & 0.15 & 0.06 & 0.00 & 0.03 & 0.15 & 0.06 & 0.00 & 0.01 & 0.036 & 0.041 \\
 & \textbf{Laion}$^\dagger$ & 0.13 & 0.05 & 0.00 & 0.00 & 0.13 & 0.05 & 0.00 & 0.02 & 0.17 & 0.07 & 0.00 & 0.01 & 0.035 & 0.040 \\
\midrule
SSA-CWA~\citep{attackbard} & \textbf{Ensemble} & 0.23 & 0.11 & 0.02 & 0.11 & 0.26 & 0.09 & 0.02 & 0.12 & 0.40 & 0.25 & 0.08 & 0.17 & 0.059 & 0.060 \\
AnyAttack~\citep{anyattack} & \textbf{Ensemble} & 0.48 & 0.24 & 0.05 & 0.28 & 0.49 & 0.29 & 0.05 & 0.19 & 0.53 & 0.32 & 0.08 & 0.31 & 0.054 & 0.057 \\
FOA-Attack~\citep{jia2025adversarial} & \textbf{Ensemble} & 0.47 & 0.31 & 0.12 & 0.44 & 0.54 & 0.40 & 0.10 & 0.45 & 0.94 & 0.78 & 0.37 & 0.87 & 0.031 & 0.036 \\
SGHA-Attack~\citep{wang2026sgha} & \textbf{Ensemble} & 0.17 & 0.07 & 0.00 & 0.02 & 0.12 & 0.03 & 0.00 & 0.03 & 0.17 & 0.07 & 0.00 & 0.01 & 0.033 & 0.039 \\
\textbf{\attackname{}}~\citep{mattack} & \textbf{Ensemble} & 0.51 & 0.34 & 0.10 & 0.40 & 0.46 & 0.30 & 0.06 & 0.40 & 0.91 & 0.71 & 0.30 & 0.85 & 0.030 & 0.036 \\
\textbf{\ours{}} (Ours) & \textbf{Ensemble} & \cellcolor[gray]{0.9}0.69 & \cellcolor[gray]{0.9}0.50 & \cellcolor[gray]{0.9}0.20 & \cellcolor[gray]{0.9}0.60 & \cellcolor[gray]{0.9}0.70 & \cellcolor[gray]{0.9}0.48 & \cellcolor[gray]{0.9}0.16 & \cellcolor[gray]{0.9}0.61 & \cellcolor[gray]{0.9}\textbf{0.99} & \cellcolor[gray]{0.9}0.88 & \cellcolor[gray]{0.9}0.48 & \cellcolor[gray]{0.9}0.97 & \cellcolor[gray]{0.9}0.038 & \cellcolor[gray]{0.9}0.044 \\
\midrule
MPCAttack~\citep{li2026multi} & \textbf{Ens.+LVLM} & 0.51 & 0.35 & 0.08 & 0.48 & 0.47 & 0.27 & 0.07 & 0.38 & 0.97 & \textbf{0.90} & \textbf{0.51} & 0.98 & 0.042 & 0.047 \\
\textbf{\ours{}} (Ours) & \textbf{Ens.+LVLM} & \cellcolor[gray]{0.9}\textbf{0.82} & \cellcolor[gray]{0.9}\textbf{0.62} & \cellcolor[gray]{0.9}\textbf{0.29} & \cellcolor[gray]{0.9}\textbf{0.73} & \cellcolor[gray]{0.9}\textbf{0.82} & \cellcolor[gray]{0.9}\textbf{0.58} & \cellcolor[gray]{0.9}\textbf{0.28} & \cellcolor[gray]{0.9}\textbf{0.75} & \cellcolor[gray]{0.9}\textbf{0.99} & \cellcolor[gray]{0.9}\textbf{0.90} & \cellcolor[gray]{0.9}0.45 & \cellcolor[gray]{0.9}\textbf{0.99} & \cellcolor[gray]{0.9}0.038 & \cellcolor[gray]{0.9}0.044 \\
\bottomrule
\end{tabular}
\end{adjustbox}
\caption{Comparison of attack methods on three black-box commercial LVLMs. $^\dagger$: pre-trained on LAION~\citep{laion5b}. Ens.+LVLM: CLIP ensemble plus InternVL3-1B and DINOv2.} \label{tab:main-compare}
\end{table*}

\section{Experiments}

\subsection{Experimental Setup} \label{sec:setup}

\paragraph{Metrics.} We adopt the evaluation protocol of \attackname{}, reporting the \emph{Attack Success Rate} (ASR) via \textit{GPTScore} and the \emph{Keywords Matching Rate} (KMR) at three thresholds $\{0.25,0.5,1.0\}$, denoted as KMR$_a$, KMR$_b$, and KMR$_c$~\citep{mattack}. KMR measures semantic alignment using human-annotated keywords, considering a match successful if the rate exceeds threshold $x$. GPT-6.1 Sol serves as the judge for both metrics.

\paragraph{Victim models and dataset.} We evaluate state-of-the-art commercial LVLMs: GPT-6 Astra, GPT-6.1 Sol, GPT-6 Luna, Claude Sonnet 5.5, Grok 4.7, Gemini 3.8 Flash~\citep{gemini}, and the vision-reasoning model GPT-o3, as well as the open-source Qwen3.5-9B and InternVL3.5-8B; further studies in the appendix cover additional models from the GPT, Claude, and Gemini families. Following SSA-CWA~\citep{attackbard} and \attackname{}~\citep{mattack}, clean (source) images and target images are drawn from the \emph{NIPS 2017 Adversarial Attacks and Defenses Competition} dataset~\citep{nips-2017-defense-against-adversarial-attack}, and \textit{\textbf{auxiliary sets} are retrieved from the COCO training set~\citep{coco} using CLIP-B/16 embedding similarity}. The \textit{milder target transformation} includes random resized crop ($[0.9,1.0]$), random horizontal flip ($p=0.5$), and random rotation ($\pm15^\circ$).

\paragraph{Hyperparameters.}
Unless noted, perturbations are bounded by $\ell_\infty$ with $\epsilon=16$ and optimized for 300 steps.
For \ours{}, $\alpha=1.275$, $\beta_1=0.9, \beta_2=0.99$ for momentum, $K=10$, $P=2$, and $\lambda=0.3$ for MCA and ATA (protocol details in \cref{appendix:details}). The surrogate ensemble is PE$^{+}$: CLIP$^\dag$-G/14, CLIP-B/16, CLIP-B/32, and CLIP$^\dag$-B/32, selected in \cref{appendix:pe}.

\begin{table}[t]
\centering
\footnotesize
\setlength{\tabcolsep}{1.5pt}
\begin{adjustbox}{max width=\linewidth,center}
\begin{tabular}{@{}l|cccc|cccc@{}}
\toprule
\multirow{2}{*}{Method} &
\multicolumn{4}{c|}{Qwen3.5-9B} &
\multicolumn{4}{c}{InternVL3.5-8B} \\
\cmidrule(lr){2-5}\cmidrule(lr){6-9}
& \kh{KMR$_a$} & \kh{KMR$_b$} & \kh{KMR$_c$} & \kh{ASR}
& \kh{KMR$_a$} & \kh{KMR$_b$} & \kh{KMR$_c$} & \kh{ASR} \\
\midrule
AttackVLM & 0.12 & 0.06 & 0.00 & 0.02 & 0.13 & 0.06 & 0.01 & 0.03 \\
SSA-CWA & 0.56 & 0.44 & 0.12 & 0.49 & 0.43 & 0.30 & 0.05 & 0.32 \\
AnyAttack & 0.64 & 0.45 & 0.08 & 0.46 & 0.66 & 0.40 & 0.13 & 0.46 \\
\midrule
FOA-Attack & 0.97 & 0.86 & 0.41 & 0.98 & \textbf{0.99} & 0.86 & 0.48 & 0.95 \\
\attackname{} & 0.96 & 0.81 & 0.40 & 0.94 & 0.95 & 0.78 & 0.39 & 0.91 \\
\ours{} & \cellcolor[gray]{0.9}\textbf{0.98} & \cellcolor[gray]{0.9}\textbf{0.88} & \cellcolor[gray]{0.9}\textbf{0.53} & \cellcolor[gray]{0.9}\textbf{1.00} & \cellcolor[gray]{0.9}\textbf{0.99} & \cellcolor[gray]{0.9}\textbf{0.91} & \cellcolor[gray]{0.9}\textbf{0.49} & \cellcolor[gray]{0.9}\textbf{0.99} \\
\bottomrule
\end{tabular}
\end{adjustbox}
\caption{Comparison on open-source LVLMs (Qwen3.5-9B and InternVL3.5-8B). Higher KMR$_{a/b/c}$ and ASR are better.}
\label{tab:qwen-llava-compare}
\par\medskip
\begin{adjustbox}{max width=0.82\linewidth,center}
\begin{tabular}{lcccc}
\toprule
Model & KMR$_a$ & KMR$_b$ & KMR$_c$ & ASR \\
\midrule
GPT-o3 (o3-2025-04-16) & 0.91 & 0.71 & 0.23 & 0.98 \\
\bottomrule
\end{tabular}
\end{adjustbox}
\caption{Results of \ours{} on the vision-reasoning model GPT-o3.}
\label{tab:reasoning}
\end{table}

\begin{figure}[t]
    \centering
    \includegraphics[width=\linewidth,trim=0 0 376 0,clip]{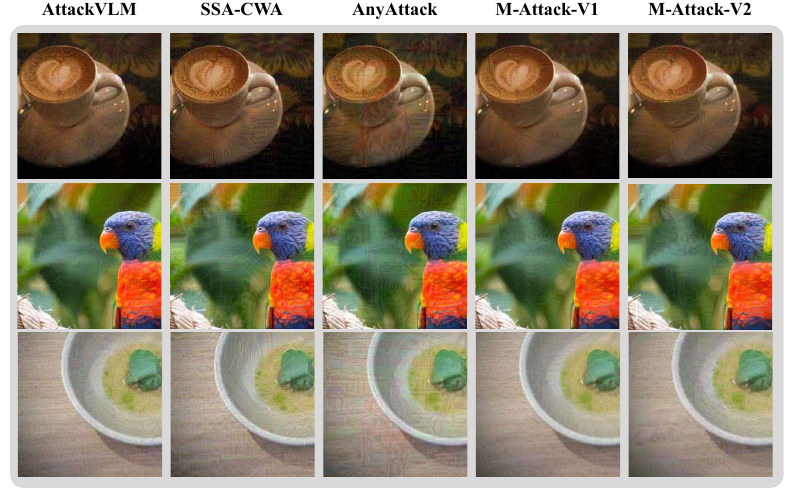}
    \caption{Adversarial samples of different attacks under $\epsilon=8$. More examples are in \cref{appendix:vis}.}
    \label{fig:epsilon 8}
\end{figure}

\subsection{Main Results} \label{sec:main_compare}

\paragraph{Transferability across LVLMs.} Tab.~\ref{tab:main-compare} reports the main comparison on frontier commercial LVLMs, and Tab.~\ref{tab:qwen-llava-compare} extends the evaluation to open-source LVLMs. In the upper block of Tab.~\ref{tab:main-compare}, \ours{} consistently outperforms prior transfer attacks, including \attackname{}. \ours{} reaches 60\% ASR on GPT-6 Astra and 61\% on GPT-6.1 Sol, against at most 45\% for prior attacks. On Claude Sonnet 5.5, where the strongest prior attacks already reach 85--87\% ASR, \ours{} still raises ASR to 97\% and KMR$_c$ from 0.37 to 0.48. GPT-6 Luna, Grok 4.7, and Gemini 3.8 Flash (\cref{tab:grok-gemini}) show the same ranking. The stronger optimization slightly increases the $\ell_1$ and $\ell_2$ norms because \ours{} makes fuller use of the allowed $\ell_\infty$ ball. Fig.~\ref{fig:epsilon 8} shows representative adversarial examples, and the human perceptual study in Sec.~\ref{subsec:human_perceptual} confirms that \ours{} remains visually comparable to \attackname{} while outperforming other baselines. PE$^+$ contains only CLIP encoders; MPCAttack~\citep{li2026multi} also uses InternVL3-1B and DINOv2. With PE$^+$ plus these two models (Ens.+LVLM), \ours{} achieves higher ASR than MPCAttack on all three victims in Tab.~\ref{tab:main-compare}.

\paragraph{Robustness against vision-reasoning models.}
Reasoning in text modality does not directly alter the visual features delivered by the vision backbone. We therefore evaluate \ours{} against GPT-o3, a model enhanced with visual reasoning capabilities. As shown in Tab.~\ref{tab:reasoning}, visual reasoning adds little robustness, and GPT-o3 remains highly vulnerable. Fig.~\ref{fig:response} in \cref{appendix:vis} provides representative responses: GPT-o3 can notice suspicious visual artifacts in some images, yet the final answer still follows the injected target semantics. This suggests that current visual reasoning can describe low-level anomalies without reliably rejecting adversarial semantic alignment.

\begin{table}[t]
  \centering
    \footnotesize
  \setlength{\tabcolsep}{1.5pt}
  \renewcommand{\arraystretch}{1.05}
  \begin{adjustbox}{max width=\columnwidth,center}
    \begin{tabular}{c c c | c c c c | c c c c }
      \toprule
        \multicolumn{3}{c|}{\textbf{Component}} &
        \multicolumn{4}{c|}{\textbf{GPT-6 Luna}} &
        \multicolumn{4}{c}{\textbf{GPT-6 Astra}}             \\
      \cmidrule(lr){1-3}\cmidrule(lr){4-7}\cmidrule(lr){8-11}
        \kh{MCA} & \kh{ATA} & \kh{PM} &
        \kh{KMR$_a$} & \kh{KMR$_b$} & \kh{KMR$_c$} & \kh{ASR} &
        \kh{KMR$_a$} & \kh{KMR$_b$} & \kh{KMR$_c$} & \kh{ASR}  \\
      \midrule
      &       &       &
        0.53 &
        0.36 &
        0.15 &
        0.58 &
        0.79 &
        0.56 &
        0.21 &
        0.68 \\
\midrule
\ding{55} &       &        &
        \val{0.46}{$\downarrow0.07$} &
        \val{0.30}{$\downarrow0.06$} &
        \val{0.08}{$\downarrow0.07$} &
        \val{0.44}{$\downarrow0.14$} &
        \val{0.64}{$\downarrow0.15$} &
        \val{0.39}{$\downarrow0.17$} &
        \val{0.15}{$\downarrow0.06$} &
        \val{0.57}{$\downarrow0.11$} \\
      & \ding{55} &        &
        \val{0.43}{$\downarrow0.10$} &
        \val{0.27}{$\downarrow0.09$} &
        \val{0.08}{$\downarrow0.07$} &
        \val{0.46}{$\downarrow0.12$} &
        \val{0.75}{$\downarrow0.04$} &
        \val{0.51}{$\downarrow0.05$} &
        \val{0.17}{$\downarrow0.04$} &
        \val{0.62}{$\downarrow0.06$} \\
\ding{55} & \ding{55} &    &
        \val{0.42}{$\downarrow0.11$} &
        \val{0.22}{$\downarrow0.14$} &
        \val{0.05}{$\downarrow0.10$} &
        \val{0.32}{$\downarrow0.26$} &
        \val{0.61}{$\downarrow0.18$} &
        \val{0.43}{$\downarrow0.13$} &
        \val{0.15}{$\downarrow0.06$} &
        \val{0.49}{$\downarrow0.19$} \\
            &       & \ding{55} &
        \val{0.47}{$\downarrow0.06$} &
        \val{0.28}{$\downarrow0.08$} &
        \val{0.09}{$\downarrow0.06$} &
        \val{0.46}{$\downarrow0.12$} &
        \val{0.73}{$\downarrow0.06$} &
        \val{0.58}{$\uparrow0.02$} &
        \val{0.22}{$\uparrow0.01$} &
        \val{0.70}{$\uparrow0.02$} \\
      \bottomrule
    \end{tabular}%
    \end{adjustbox}
\caption{Removing each component (\ding{55}). Small numbers give the change from the full model (first row).}
  \label{tab:component_ablation_delta}
\end{table}

\subsection{Ablation Studies} \label{sec:ablation}

\paragraph{Module contributions.}
Tab.~\ref{tab:component_ablation_delta} isolates the effect of each module beyond PE$^+$. MCA and ATA are the two main sources of improvement. Removing MCA consistently weakens ASR and KMR because the update returns to a single-view crop estimate with stronger ViT-induced fluctuations. Removing ATA particularly hurts GPT-6 Luna KMR$_b$, indicating that target-side reference coverage is important for semantically faithful transfer, not only for attack success. Removing both MCA and ATA produces the largest degradation on GPT-6 Luna, confirming that source-side crop consistency and target-side reference coverage address complementary instability sources. PM has a smaller marginal effect on GPT-6 Astra, consistent with its role as a low-cost temporal accumulator: it strengthens recurring directions but does not replace the MCA/ATA estimators. This pattern suggests that PE$^+$ provides useful architectural diversity, while MCA and ATA turn that diversity into stable transferable updates.

\begin{figure}[t]
    \centering
    \includegraphics[width=0.93\linewidth,trim=0 20 0 0,clip]{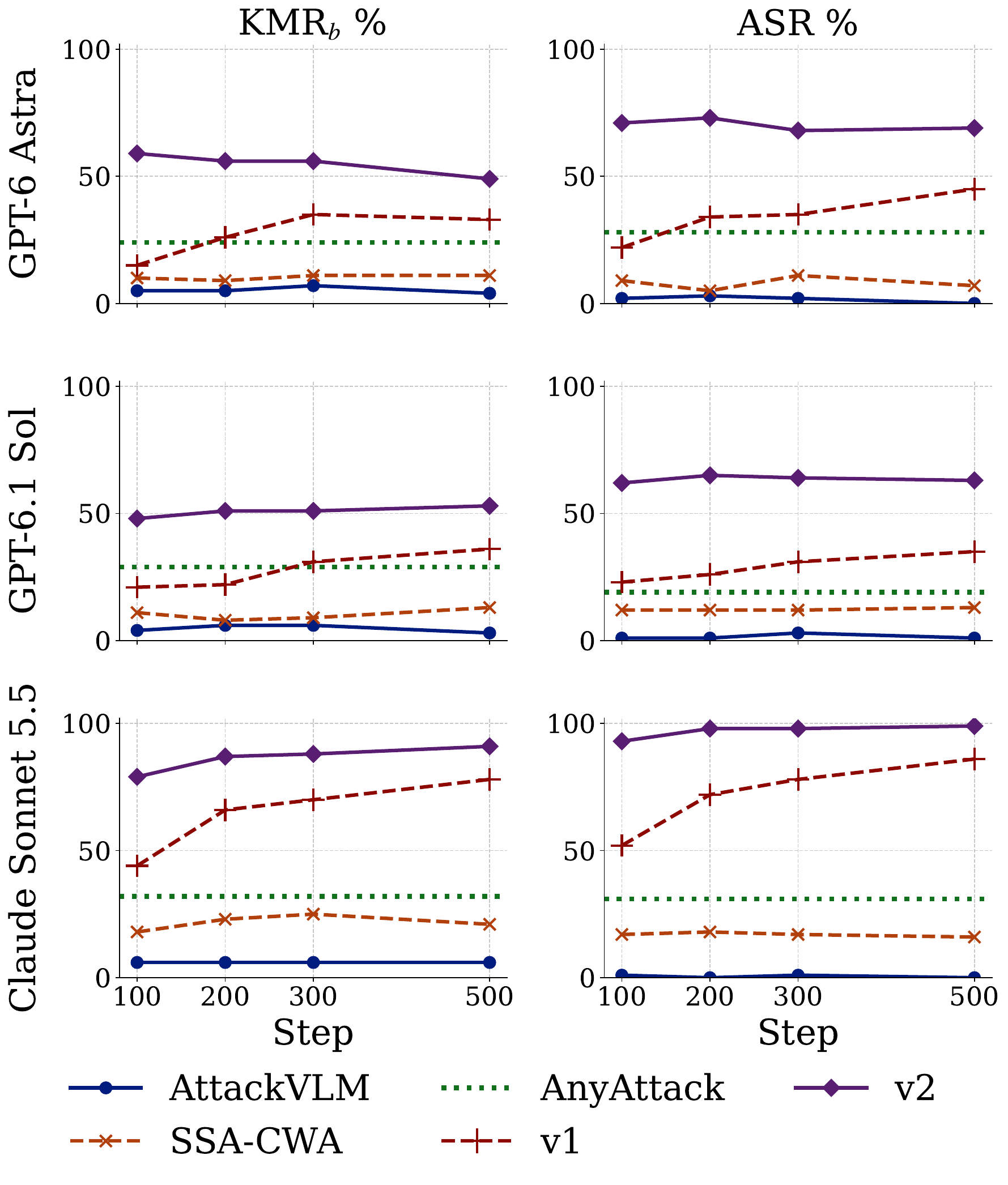}
    \caption{Comparison under different step budgets.}
    \label{fig:ablation_step}
\end{figure}

\paragraph{Step budget.}
Fig.~\ref{fig:ablation_step} compares performance under varying optimization budgets. With 100 steps, our method already exceeds \attackname{} at 500 steps on all three victims, while \attackname{} drops notably at 100 and 200 steps. Our method stays stable due to a more coherent optimization trajectory than \attackname{}, which is more sensitive to random cropping and aggressive target transformations.

\section{Conclusion}
We revisit transfer-based targeted attacks on LVLMs through local-matching instability: highly overlapping source crops can produce nearly orthogonal gradients, while target-side transformations can drift away from the intended semantics. \ours{} addresses this bottleneck with MCA, which recovers crop-consistent gradients, ATA, which expands semantic reference coverage, PM, which accumulates recurring patch-level signals, and PE$^+$, which selects patch-diverse surrogate encoders. Experiments on six commercial LVLMs and the open-source Qwen3.5-9B and InternVL3.5-8B show consistent gains under realistic perturbation budgets.

\section*{Limitations}
Our evaluation follows the protocol of \attackname{}: the main results use 100 source--target pairs from the NIPS 2017 competition set, with a 1K-image check in \cref{appendix:1k}. Commercial LVLMs are evaluated through their public APIs at a fixed point in time; providers update these models, so absolute success rates may shift for later versions. ASR relies on an LLM judge (GPTScore) and KMR on human-annotated keywords, both of which approximate whether a response carries the target semantics. GPT-6 Astra and GPT-6.1 Sol remain the hardest targets (60--61\% ASR), and at $\epsilon=4$ all evaluated transfer attacks, including ours, stay weak on them. Imperceptibility is controlled by an $\ell_\infty$ budget and checked by a small human study rather than by perceptual constraints in the optimization. Finally, ATA requires a retrieval database for auxiliary images, and MCA multiplies the number of surrogate forward and backward passes per iteration.

\section*{Ethical Considerations}
This work strengthens transfer-based black-box attacks on large vision-language models, improving the ability to stress-test real-world multimodal systems used in assistants, search, and content generation. By revealing key instabilities in local-level matching and proposing simple gradient-denoising fixes, our methods can help researchers and practitioners build more reliable defenses, develop better robustness benchmarks, and identify failure modes before deployment. At the same time, more effective attacks can be misused to bypass safety filters, induce targeted hallucinations, or manipulate model outputs in high-stakes settings. To limit misuse, we evaluate primarily on controlled benchmarks and public models, and we release code and data to support reproducibility and defense research while avoiding instructions or configurations that lower the barrier to real-world harm.

\bibliography{main}

\clearpage
\appendix

\section*{\Large{Appendix}}

\etocdepthtag.toc{mtappendix}
\etocsettagdepth{mtchapter}{none}
\etocsettagdepth{mtappendix}{subsection}
\tableofcontents

\setcounter{dbltopnumber}{1}

\section{Experimental Protocol Details} \label{appendix:details}
\begin{table*}[tp]
  \centering
  \small
    \setlength{\tabcolsep}{6pt}
  \begin{tabular}{@{}ll@{}}
    \toprule
    \textbf{Surrogate (paper notation)} & \textbf{Implementation (Hugging Face identifier)} \\
    \midrule
    CLIP$^\dag$-B/32~\citep{opanai_clip_software,laion5b}   & \texttt{laion/CLIP-ViT-B-32-laion2B-s34B-b79K}                \\
    CLIP$^\dag$-H/14~\citep{opanai_clip_software,laion5b}      & \texttt{laion/CLIP-ViT-H-14-laion2B-s32B-b79K}            \\
    CLIP-L/14~\citep{clip}          & \texttt{openai/clip-vit-large-patch14}                       \\
    CLIP$^\dag$-B/16~\citep{opanai_clip_software,laion5b}    & \texttt{laion/CLIP-ViT-B-16-laion2B-s34B-b88K}              \\
    CLIP$^\dag$-G/14~\citep{opanai_clip_software,laion5b}  & \texttt{laion/CLIP-ViT-bigG-14-laion2B-39B-b160k}            \\
    Dino-Small~\citep{dinov2}         & \texttt{facebook/dinov2-small}                               \\
    Dino-Base~\citep{dinov2}          & \texttt{facebook/dinov2-base}                                \\
    Dino-Large~\citep{dinov2}         & \texttt{facebook/dinov2-large}                               \\
    BLIP-2 (2.7B)~\citep{blip2}     & \texttt{Salesforce/blip2-opt-2.7b}                           \\
    \bottomrule
  \end{tabular}
\caption{Surrogate models and their corresponding Hugging Face identifiers.}
  \label{tab:surrogates}
\end{table*}

All experiments use seed 2023. The original runs used Ubuntu 22.04 with 6 NVIDIA RTX 4090 GPUs; the sets regenerated for the step-budget and 1K-image experiments (\attackname{}, AttackVLM, AnyAttack, and \ours{}) ran on 4 NVIDIA A100 GPUs. For baselines, we set the step size to $\alpha=0.75$ for Claude and $\alpha=1.0$ for all other models, mirroring \attackname{}. We set the ASR threshold to 0.3 following \attackname{}. The main-text victims (GPT-6 Astra, GPT-6.1 Sol, and GPT-6 Luna at reasoning effort low, Claude Sonnet 5.5 with adaptive thinking at effort low, Grok 4.7, and Gemini 3.8 Flash) are queried through their APIs with default sampling, Qwen3.5-9B and InternVL3.5-8B run locally with greedy decoding, and GPT-6.1 Sol (reasoning effort low) judges both ASR and KMR. The earlier models in the appendix were queried with temperature 0 and judged by GPT-4o. The SSA-CWA implementation reads its inputs in lexicographic file order, so each of its outputs is scored against the target it was crafted for; AnyAttack is generated on the same 100 source--target pairs. SGHA-Attack runs its official objective with $\epsilon=16$ and 300 steps, averaged over CLIP-B/16, CLIP-B/32, and CLIP-L/14; the official code attacks a single encoder, and with CLIP-B/32 alone its ASR stays at or below 2\% on all six commercial models. Its Stable Diffusion anchor images are not released, so we use the official fallback that takes the target image as the anchor. MPCAttack runs its official code and default settings ($\epsilon=16$, 300 steps), whose surrogates are CLIP-B/16, CLIP-B/32, CLIP$^\dag$-G/14, InternVL3-1B, and DINOv2-B. The Ens.+LVLM row of \ours{} in \cref{tab:main-compare,tab:grok-gemini} adds InternVL3-1B and DINOv2-B to PE$^+$ and keeps every other setting. Following SSA-CWA~\citep{attackbard} and \attackname{}~\citep{mattack}, we randomly sample 100 source--target image pairs for the main experiments; results on a larger 1K-image subset are reported in \cref{appendix:1k}. GPT-5 results in the appendix use GPT-5-thinking-low, i.e., the API reasoning effort is set to low; other reasoning budgets are reported in \cref{appendix:gpt5}. All BLIP-2 variants on Hugging Face share the same vision encoder, so we use only one. \Cref{tab:surrogates} maps the surrogate names used in this paper to their Hugging Face identifiers.

\section{Additional Details for Theoretical Analysis}
\subsection{Proof of Theorem 1} \label{sec: theory}

This section provides detailed proof of the upper bound in Eq.~(\ref{equ:bound}). For variance, we have
{\allowdisplaybreaks
\begin{align}
\operatorname{Var}(\hat{g}_K) &:= \mathbb{E}\left\|\hat{g}_K-\mu\right\|^2 \nonumber \\
&= \mathbb{E}\left\|\frac{1}{K}\sum_{k=1}^K(g_k-\mu)\right\|^2 \nonumber \\
&= \frac{1}{K^2}\sum_{k=1}^K\sum_{\ell=1}^K\mathbb{E}[(g_k-\mu)^\top(g_\ell-\mu)] \nonumber \\
&= \frac{1}{K^2} \bigg( \underbrace{\sum_{k=1}^K \mathbb{E}\|g_k - \mu\|_2^2}_{K\sigma^2} \nonumber \\
&\qquad + \underbrace{2 \sum_{1 \leq k < \ell \leq K} \mathbb{E}[\langle g_k - \mu, g_\ell - \mu \rangle]}_{\text{cross terms}} \bigg) \label{equ:expand}
\end{align}}
The diagonal part is reduced to the mean. We now provide an upper bound for the cross terms. Let
$\rho_{k \ell} = \frac{\left < g_k - \mu, g_\ell - \mu \right >}{\lVert g_k - \mu\rVert_2\lVert g_\ell - \mu\rVert_2}$ denote the pairwise residual correlation. Then,
\begin{equation}
\begin{aligned}
    &\mathbb{E}[\langle g_k-\mu,g_\ell-\mu\rangle] \\
    &\quad =\mathbb{E}\left[\rho_{k\ell}\|g_k-\mu\|_2\|g_\ell-\mu\|_2\right]. \label{equ:corss}
\end{aligned}
\end{equation}
Since all crops share the same marginal distribution, applying the Cauchy-Schwarz inequality to Eq.~(\ref{equ:corss}) yields
\begin{equation}
\begin{aligned}
    &\mathbb{E}[\langle g_k-\mu,g_\ell-\mu\rangle] \\
    &\quad \leq\mathbb{E}[\rho_{k\ell}]\sqrt{\mathbb{E}\|g_k-\mu\|_2^2}\sqrt{\mathbb{E}\|g_\ell-\mu\|_2^2}\\
    &\quad =\bar{\rho}\sigma^2,
\end{aligned}
\end{equation}
where $\overline{\rho}=\mathbb{E}[\rho_{k\ell}], k\neq \ell$. Plugging this into the double sum term yields
\begin{equation}
\begin{aligned}
    &\sum_{1\leq k<\ell\leq K}\mathbb{E}\left[\langle g_k-\mu,g_\ell-\mu\rangle\right] \\
    &\quad \leq \frac{K(K-1)}{2}\bar{\rho}\sigma^2.
\end{aligned} \label{equ:cross:new}
\end{equation}
The $\frac{K(K-1)}{2}$ appears since there are total $\frac{K(K-1)}{2}$ terms for $\sum_{k<\ell}$. Thus substituting Eq.~(\ref{equ:cross:new}) back to the cross-term part in Eq.~(\ref{equ:expand}) yields
\begin{equation}
\begin{aligned}
    \operatorname{Var}(\hat{g}_K) & \le  \frac{1}{K^2}\left ( K \sigma^2 + K(K-1)\overline{\rho}\sigma^2 \right  ) \\ &= \frac{1}{K}\sigma^2 + \frac{K-1}{K}\overline{\rho}\sigma^2.
    \end{aligned}
\end{equation}
Therefore, we have the upper bound provided in Sec.~\ref{sec:MCA}.

\subsection{Proof of Theorem 2} \label{sec:proof-drift}

We begin with the drift analysis for \attackname{}:
\begin{equation}
\begin{aligned}
&\Delta_{\text{drift}}(\mathcal{T};\mathbf{X}_\mathrm{tar}) \\
&\quad = \mathbb{E}_{\mathcal{T}\sim D_{\alpha}}[\lVert f(\mathcal{T}(\mathbf{X}_\mathrm{tar})) - f(\mathbf{X}_\mathrm{tar}) \rVert] \\
&\quad \le L \cdot \mathbb{E}_{\mathcal{T}\sim D_{\alpha}}[\lVert \mathcal{T}(\mathbf{X}_{\mathrm{tar}})-\mathbf{X}_{\mathrm{tar}} \rVert]\\
&\quad \le L\alpha .
\end{aligned}
\end{equation}
Next, we analyze the drift for \ours{} using the triangle inequality and (A1)--(A3). Writing $\mathbf{X}_p := \mathbf{X}^{(p)}_{\text{aux}}$ for brevity,
\begin{equation*}
\begin{aligned}
&\Delta_{\text{drift}}(\tilde{\mathcal T};\mathbf{X}_p) \\
&= \mathbb{E}_{\tilde{\mathcal T}\sim D_{\tilde{\alpha}}}\!\bigl[\| f(\tilde{\mathcal T}(\mathbf{X}_p)) - f(\mathbf X_{\text{tar}}) \|\bigr] \\
&\le \mathbb{E}_{\tilde{\mathcal T}}\!\bigl[\| f(\tilde{\mathcal T}(\mathbf{X}_p))-f(\mathbf{X}_p)\| \\
&\qquad\quad + \| f(\mathbf{X}_p)-f(\mathbf X_{\text{tar}}) \| \bigr] \\
&= \mathbb{E}_{\tilde{\mathcal T}}\!\bigl[\| f(\tilde{\mathcal T}(\mathbf{X}_p))-f(\mathbf{X}_p) \| \bigr] \\
&\qquad\quad + \mathbb{E}\!\bigl[\| f(\mathbf{X}_p)-f(\mathbf X_{\text{tar}}) \| \bigr] \\
&\le L\,\mathbb{E}_{\tilde{\mathcal T}}\!\bigl[\| \tilde{\mathcal T}(\mathbf{X}_p)-\mathbf{X}_p \| \bigr] + \delta \\
&\le L\tilde{\alpha} + \delta.
\end{aligned}
\end{equation*}
Thus, we have completed the proof of Theorem~\ref{thm:drift}.

\subsection{Justification for Assumptions} \label{appendix:justification}

Assumption (A2) is derived from the retrieval mechanism for auxiliary data. Specifically, $\mathbf{X}_{\text{aux}}^{(p)}$ represents the $p$-th closest embedding to the target $\mathbf{X}_{\text{tar}}$ from a database $\mathcal{D}$, defined explicitly by:
\begin{equation}
\begin{aligned}
\mathbf{X}_{\text{aux}}^{(p)} \in \arg\mathrm{top}_P \Big\{ & \mathbf{X} \in \mathcal{D}: \\
& \frac{f(\mathbf{X})^\top f(\mathbf{X}_{\text{tar}})}{|f(\mathbf{X})||f(\mathbf{X}_{\text{tar}})|} \Big\},
\end{aligned}
\end{equation}
where $\mathrm{top}_P$ denotes selecting the top-$P$ nearest neighbors according to cosine similarity. Given that embeddings $f(\mathbf{X})$ are typically normalized, semantic closeness naturally bounds the expected distance between $f(\mathbf{X}_{\text{aux}}^{(p)})$ and $f(\mathbf{X}_{\text{tar}})$ by $\delta$, thus validating (A2). Under normalized embeddings, $\delta$ can be estimated by $2\big(1-f(\mathbf{X}_{\rm aux}^{(P)})^\top f(\mathbf{X}_{\rm tar})\big)$.

\section{Optimizer Variants} \label{appendix:alg}
\begin{algbox}
\algcaption{\ours{} (Adam variant)}
\label{alg:mattack_v2_adam}
\footnotesize
\begin{algbody}
\begin{algorithmic}[1]
\REQUIRE $\mathbf{X}_{\mathrm{clean}}$, $\mathbf{X}_{\mathrm{tar}}$, auxiliary set $\mathcal{A}$, ensemble $\Phi^+$, iterations $n$, step size $\alpha$, budget $\epsilon$, Adam $\beta_1,\beta_2,\eta$, crops $K$, auxiliary weight $\lambda$.
\ENSURE Adversarial image $\mathbf{X}_{\mathrm{adv}}$.
\STATE $\mathbf{X}_{\mathrm{adv}}\gets\mathbf{X}_{\mathrm{clean}}$; $m\gets0$; $v\gets0$.
\FOR{$i=1$ to $n$}
    \STATE Draw $\{\mathcal{T}_k\}_{k=1}^{K}\sim\mathcal{D}$; set $g\gets0$.
    \FOR{$k=1$ to $K$}
        \STATE Draw $\{\tilde{\mathcal{T}}_p\}_{p=0}^{P}\sim\tilde{\mathcal{D}}$.
        \FOR{$j=1$ to $m$}
            \STATE Compute $y_0$ from the primary target and $y_p$ from each auxiliary target.
            \STATE Accumulate $\hat{\mathcal{L}}_k$ using the primary and auxiliary alignment losses.
            \STATE $g\gets g+\frac{1}{Km}\nabla_{\mathbf{X}_{\mathrm{adv}}}\hat{\mathcal{L}}_k$.
        \ENDFOR
    \ENDFOR
    \STATE $m\gets\beta_1m+(1-\beta_1)g$; $v\gets\beta_2v+(1-\beta_2)g^{\odot2}$.
    \STATE $\hat m\gets m/(1-\beta_1^i)$; $\hat v\gets v/(1-\beta_2^i)$.
    \STATE $\mathbf{X}_{\mathrm{adv}}\gets\mathrm{clip}_{\mathbf{X}_{\mathrm{clean}},\epsilon}(\mathbf{X}_{\mathrm{adv}}+\alpha\,\hat m/(\sqrt{\hat v}+\eta))$.
\ENDFOR
\RETURN $\mathbf{X}_{\mathrm{adv}}$.
\end{algorithmic}
\end{algbody}
\algendrule
\end{algbox}
\begin{algbox}
\algcaption{\ours{} (MI-FGSM variant)}
\label{alg:mattack_v2_mifgsm}
\footnotesize
\begin{algbody}
\begin{algorithmic}[1]
\REQUIRE $\mathbf{X}_{\mathrm{clean}}$, $\mathbf{X}_{\mathrm{tar}}$, auxiliary set $\mathcal{A}$, ensemble $\Phi^+$, iterations $n$, step size $\alpha$, budget $\epsilon$, momentum decay $\gamma$, crops $K$, auxiliary weight $\lambda$.
\ENSURE Adversarial image $\mathbf{X}_{\mathrm{adv}}$.
\STATE $\mathbf{X}_{\mathrm{adv}}\gets\mathbf{X}_{\mathrm{clean}}$; $\mu\gets0$.
\FOR{$i=1$ to $n$}
    \STATE Estimate $g$ with the MCA/ATA crop loop in Alg.~\ref{alg:mattack_v2}.
    \STATE $\mu\gets\gamma\mu+g/\lVert g\rVert_1$.
    \STATE $\mathbf{X}_{\mathrm{adv}}\gets\mathrm{clip}_{\mathbf{X}_{\mathrm{clean}},\epsilon}(\mathbf{X}_{\mathrm{adv}}+\alpha\,\mathrm{sign}(\mu))$.
\ENDFOR
\RETURN $\mathbf{X}_{\mathrm{adv}}$.
\end{algorithmic}
\end{algbody}
\algendrule
\end{algbox}
\begin{figure*}[tp]
  \centering
  \includegraphics[width=0.8\textwidth,trim=0 0 64 0,clip]{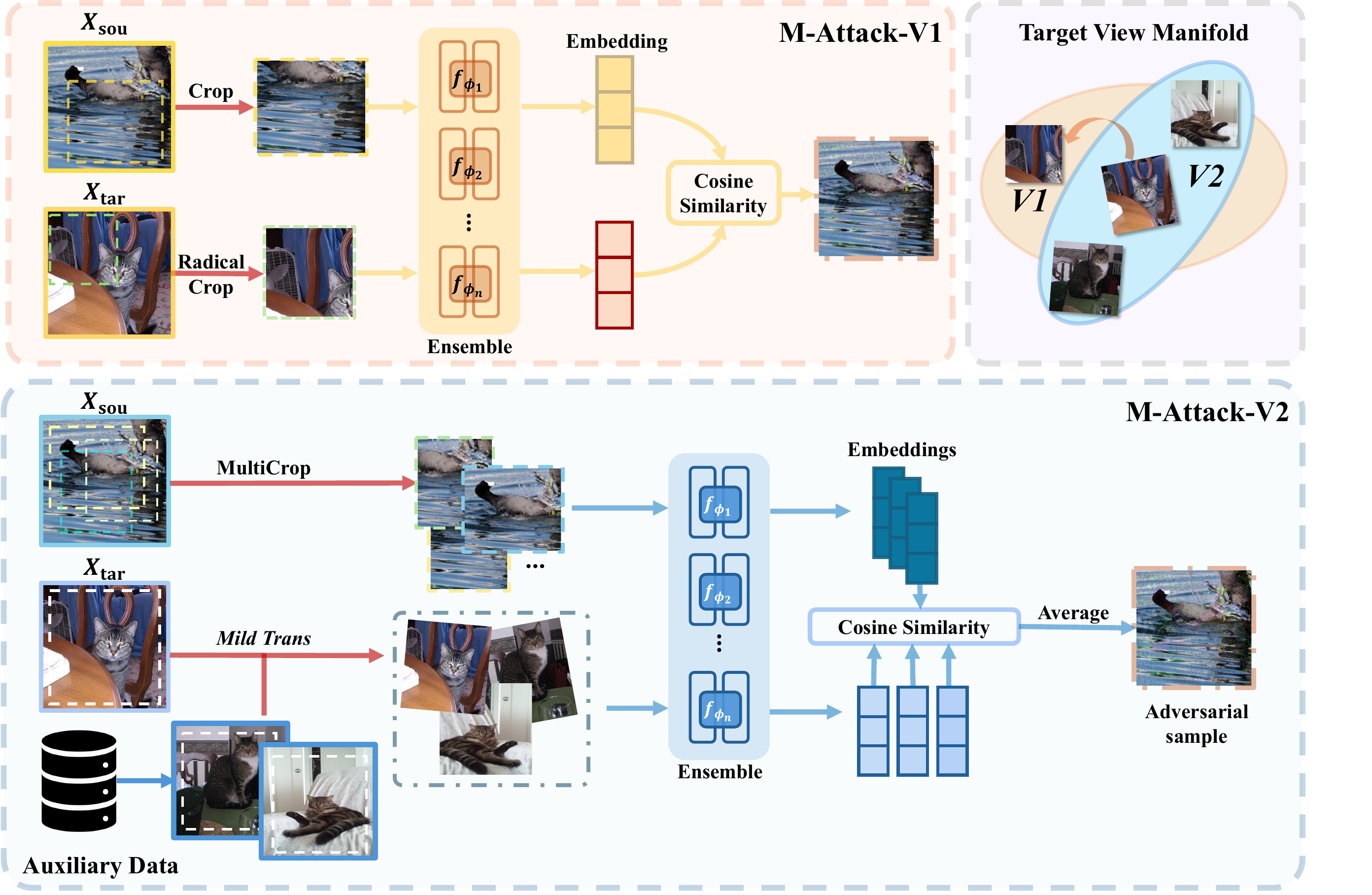}
  \caption{Overall procedure of \ours{}, contrasting the local-matching pipeline of \attackname{} with the MCA/ATA-enhanced pipeline of \ours{}. \attackname{} aligns one source crop to aggressively transformed target views, while \ours{} aggregates multiple source crops and mildly transformed auxiliary anchors, yielding smoother gradients for the final adversarial update.}
  \label{fig:full compare}
\end{figure*}
\begin{table*}[tp]
\centering
\renewcommand\arraystretch{1.16}
\begin{adjustbox}{width=0.9\textwidth,center}
\small
\setlength{\tabcolsep}{2pt}
\begin{tabular}{c|c|cccc|cccc|cccc}
\toprule
\multirow{2}{*}{Method} & \multirow{2}{*}{Model} &
\multicolumn{4}{c|}{\textbf{GPT-4o}} &
\multicolumn{4}{c|}{\textbf{Claude 3.7-extended}} &
\multicolumn{4}{c}{\textbf{Gemini 2.5-Pro}} \\
\cmidrule{3-14}
& & $\text{KMR}_a$ & $\text{KMR}_b$ & $\text{KMR}_c$ & ASR
  & $\text{KMR}_a$ & $\text{KMR}_b$ & $\text{KMR}_c$ & ASR
  & $\text{KMR}_a$ & $\text{KMR}_b$ & $\text{KMR}_c$ & ASR \\
\midrule
\ours{}-ADAM (Ours) & Ensemble & 0.91 & 0.78 & 0.40 & 0.99 & 0.56 & 0.32 & 0.11 & 0.67 & 0.87 & 0.72 & 0.22 & 0.97 \\
\ours{}-FGSM        & Ensemble & 0.85 & 0.64 & 0.19 & 0.98 & 0.40 & 0.26 & 0.08 & 0.46 & 0.83 & 0.65 & 0.17 & 0.90 \\
\ours{}-MIFGSM      & Ensemble & 0.90 & 0.66 & 0.23 & 0.96 & 0.45 & 0.30 & 0.07 & 0.57 & 0.84 & 0.64 & 0.15 & 0.87 \\
\bottomrule
\end{tabular}
\end{adjustbox}
\caption{Ablation study of \ours{} under different optimizer/attack variants.}
\label{tab:ablation-ours}
\end{table*}

Alg.~\ref{alg:mattack_v2_adam} and Alg.~\ref{alg:mattack_v2_mifgsm} spell out the optimizer variants for the update step in Alg.~\ref{alg:mattack_v2}. \Cref{tab:ablation-ours} compares them: \ours{} remains effective under FGSM-style updates, and the Adam variant used in the main paper performs best.

\section{Surrogate Model Selection} \label{appendix:pe}
\begin{figure*}[tp]
    \centering
    \includegraphics[width=0.93\textwidth]{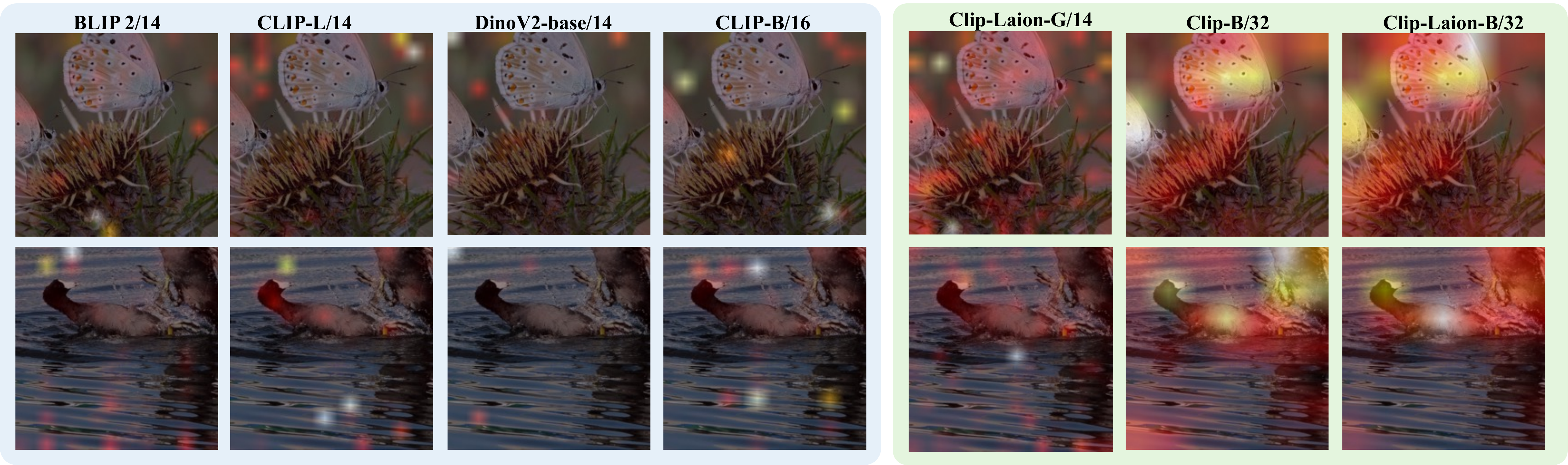}
    \caption{Comparison of two types of attention maps. Left: attention map that sparsely separates into different regions; right: attention map that focuses on the main object.}
    \label{fig:attention map}
\end{figure*}
\begin{table*}[tp]
    \centering
        \begin{adjustbox}{max width=0.98\textwidth,center}
    \small
    \setlength{\tabcolsep}{2pt}
    \begin{tabular}{lcccccccccccccc}
    \toprule
    Surrogate & C\textminus L/14 & C$^\dag$\textminus L/14 & D\textminus S/14 & D\textminus B/14 & D\textminus L/14 & C\textminus B/16 & C$^\dag$\textminus B/16 & C\textminus B/32 & C$^\dag$\textminus B/32 & BLIP2 & Avg/14 & Avg/16 & Avg/32 & Avg/All \\
    \midrule
    C\textminus L/14          & N/A & 0.40 & 0.10 & 0.13 & 0.12 & 0.45 & 0.40 & 0.34 & 0.24 & 0.48 & 0.25 & 0.42 & 0.29 & 0.30 \\
    \cellcolor[gray]{0.9}C$^\dag$\textminus \cellcolor[gray]{0.9}L/14
    & \cellcolor[gray]{0.9}0.44 & \cellcolor[gray]{0.9}N/A & \cellcolor[gray]{0.9}0.24 & \cellcolor[gray]{0.9}0.24 & \cellcolor[gray]{0.9}0.21 & \cellcolor[gray]{0.9}0.55 & \cellcolor[gray]{0.9}0.57 & \cellcolor[gray]{0.9}0.37 & \cellcolor[gray]{0.9}0.33 & \cellcolor[gray]{0.9}0.61 & \cellcolor[gray]{0.9}0.35 & \cellcolor[gray]{0.9}0.56 & \cellcolor[gray]{0.9}0.35 & \cellcolor[gray]{0.9}0.39 \\
    D\textminus S/14          & 0.25 & 0.39 & N/A & 0.45 & 0.38 & 0.41 & 0.45 & 0.32 & 0.25 & 0.46 & 0.39 & 0.43 & 0.28 & 0.37 \\
    D\textminus B/14          & 0.29 & 0.36 & 0.33 & N/A & 0.51 & 0.37 & 0.39 & 0.31 & 0.23 & 0.47 & 0.39 & 0.38 & 0.27 & 0.36 \\
    D\textminus L/14          & 0.26 & 0.31 & 0.12 & 0.32 & N/A & 0.31 & 0.34 & 0.30 & 0.21 & 0.42 & 0.29 & 0.33 & 0.26 & 0.29 \\
    \cellcolor[gray]{0.9}C\textminus \cellcolor[gray]{0.9}B/16
    & \cellcolor[gray]{0.9}0.44 & \cellcolor[gray]{0.9}0.43 & \cellcolor[gray]{0.9}0.21 & \cellcolor[gray]{0.9}0.18 & \cellcolor[gray]{0.9}0.13 & \cellcolor[gray]{0.9}N/A & \cellcolor[gray]{0.9}0.53 & \cellcolor[gray]{0.9}0.37 & \cellcolor[gray]{0.9}0.27 & \cellcolor[gray]{0.9}0.51 & \cellcolor[gray]{0.9}0.32 & \cellcolor[gray]{0.9}0.53 & \cellcolor[gray]{0.9}0.32 & \cellcolor[gray]{0.9}0.34 \\
    C$^\dag$\textminus B/16 & 0.43 & 0.51 & 0.22 & 0.21 & 0.15 & 0.57 & N/A & 0.39 & 0.34 & 0.52 & 0.34 & 0.57 & 0.36 & 0.37 \\
    \cellcolor[gray]{0.9}C\textminus \cellcolor[gray]{0.9}B/32
    & \cellcolor[gray]{0.9}0.37 & \cellcolor[gray]{0.9}0.43 & \cellcolor[gray]{0.9}0.21 & \cellcolor[gray]{0.9}0.11 & \cellcolor[gray]{0.9}0.09 & \cellcolor[gray]{0.9}0.55 & \cellcolor[gray]{0.9}0.53 & \cellcolor[gray]{0.9}N/A & \cellcolor[gray]{0.9}0.49 & \cellcolor[gray]{0.9}0.46 & \cellcolor[gray]{0.9}0.28 & \cellcolor[gray]{0.9}0.54 & \cellcolor[gray]{0.9}0.49 & \cellcolor[gray]{0.9}0.36 \\
    \cellcolor[gray]{0.9}C$^\dag$\textminus \cellcolor[gray]{0.9}B/32
    & \cellcolor[gray]{0.9}0.31 & \cellcolor[gray]{0.9}0.49 & \cellcolor[gray]{0.9}0.27 & \cellcolor[gray]{0.9}0.18 & \cellcolor[gray]{0.9}0.12 & \cellcolor[gray]{0.9}0.53 & \cellcolor[gray]{0.9}0.61 & \cellcolor[gray]{0.9}0.58 & \cellcolor[gray]{0.9}N/A & \cellcolor[gray]{0.9}0.50 & \cellcolor[gray]{0.9}0.31 & \cellcolor[gray]{0.9}0.57 & \cellcolor[gray]{0.9}0.58 & \cellcolor[gray]{0.9}0.40 \\
    BLIP2                    & 0.39 & 0.43 & 0.15 & 0.20 & 0.26 & 0.45 & 0.43 & 0.33 & 0.25 & N/A & 0.29 & 0.44 & 0.29 & 0.32 \\
    \bottomrule
    \end{tabular}
    \end{adjustbox}
\caption{Comparison of embedding transferability over 1K images. MCA/ATA excluded to show standalone performance. C/D = CLIP/DinoV2. Gray denotes selected models.}
    \label{tab:embedding trans}
\end{table*}
\begin{table*}[tp]
\centering
\renewcommand\arraystretch{1.0}
\begin{adjustbox}{max width=0.9\textwidth,center}
\footnotesize
\setlength{\tabcolsep}{1.5pt}
\begin{tabular}{c|c|cccc|cccc|cccc}
\toprule
\multirow{2}{*}{Variant} & \multirow{2}{*}{Surrogate Set (2 models)} &
\multicolumn{4}{c|}{\textbf{GPT-4o}} &
\multicolumn{4}{c|}{\textbf{Claude 3.7-extended}} &
\multicolumn{4}{c}{\textbf{Gemini 2.5-Pro}} \\
\cmidrule(lr){3-6}\cmidrule(lr){7-10}\cmidrule(lr){11-14}
& & $\text{KMR}_a$ & $\text{KMR}_b$ & $\text{KMR}_c$ & ASR
  & $\text{KMR}_a$ & $\text{KMR}_b$ & $\text{KMR}_c$ & ASR
  & $\text{KMR}_a$ & $\text{KMR}_b$ & $\text{KMR}_c$ & ASR \\
\midrule
Pair\(_1\) & {Dino-B, Dino-S} &
  \underline{0.84} & 0.57 & 0.15 & 0.91 &
  0.09 & 0.04 & 0.00 & 0.05 &
  \textbf{0.84} & 0.53 & 0.11 & 0.81 \\
Pair\(_2\) & {L16, B/16} &
  \textbf{0.86} & \textbf{0.69} & \underline{0.21} & \textbf{0.96} &
  \underline{0.16} & \underline{0.10} & \underline{0.01} & \underline{0.16} &
  \textbf{0.84} & \underline{0.59} & \underline{0.15} & \underline{0.91} \\
Pair\(_3\) & {L32, B/32} &
  0.76 & 0.52 & 0.13 & 0.79 &
  \textbf{0.46} & \textbf{0.29} & \textbf{0.06} & \textbf{0.70} &
  0.58 & 0.37 & 0.07 & 0.59 \\
Pair\(_4\) & {G/14, L14} &
  \textbf{0.86} & \underline{0.61} & \textbf{0.24} & \underline{0.94} &
  0.07 & 0.02 & 0.00 & 0.06 &
  \underline{0.82} & \textbf{0.64} & \textbf{0.23} & \textbf{0.92} \\
\bottomrule
\end{tabular}
\end{adjustbox}
\caption{Ablation on two-model surrogate sets. Bold numbers are the best in each column; underlined numbers are the second-best.}
\label{tab:pair_ensemble_ablation}
\vspace{0.8em}
\renewcommand\arraystretch{1.0}
\begin{adjustbox}{max width=0.9\textwidth,center}
\footnotesize
\setlength{\tabcolsep}{1.5pt}
\begin{tabular}{c|c|cccc|cccc|cccc}
\toprule
\multirow{2}{*}{Variant} & \multirow{2}{*}{Surrogate Set} &
\multicolumn{4}{c|}{\textbf{GPT-4o}} &
\multicolumn{4}{c|}{\textbf{Claude 3.7-extended}} &
\multicolumn{4}{c}{\textbf{Gemini 2.5-Pro}} \\
\cmidrule(lr){3-6}\cmidrule(lr){7-10}\cmidrule(lr){11-14}
& & $\text{KMR}_a$ & $\text{KMR}_b$ & $\text{KMR}_c$ & ASR
  & $\text{KMR}_a$ & $\text{KMR}_b$ & $\text{KMR}_c$ & ASR
  & $\text{KMR}_a$ & $\text{KMR}_b$ & $\text{KMR}_c$ & ASR \\
\midrule
PE\(_1\) & {B/16, B/32, L32, L16} &
  0.87 & 0.65 & 0.26 & \textbf{0.99} &
  0.54 & 0.32 & 0.07 & 0.68 &
  0.80 & 0.57 & 0.16 & 0.90 \\
PE\(_2\) & {Dino-B, B/32, L32, G/14} &
  0.87 & 0.69 & 0.28 & 0.97 &
  0.56 & 0.37 & 0.09 & 0.65 &
  \textbf{0.88} & 0.71 & 0.22 & 0.93 \\
PE\(_3\) & {L16, B/32, L32, G/14} &
  0.85 & 0.65 & 0.23 & \textbf{0.99} &
  \textbf{0.57} & \underline{0.40} & 0.09 & \textbf{0.73} &
  0.84 & 0.61 & 0.19 & 0.93 \\
PE\(_4\) & {B/16, B/32, L32, Dino-B} &
  0.89 & 0.67 & 0.19 & 0.98 &
  0.55 & \textbf{0.41} & 0.07 & 0.63 &
  0.87 & 0.67 & \textbf{0.23} & 0.96 \\
PE\(_5\) & {B/16, B/32, L32, Dino-S} &
  0.90 & 0.72 & 0.25 & 0.97 &
  0.48 & 0.33 & 0.08 & 0.59 &
  0.83 & 0.63 & 0.17 & 0.90 \\
\midrule
\cellcolor[gray]{0.9}\textbf{PE\(^{+}\) (Ours)} &
  \cellcolor[gray]{0.9}{B/16, B/32, L32, G/14} &
  \cellcolor[gray]{0.9}\textbf{0.91} & \cellcolor[gray]{0.9}\textbf{0.78} & \cellcolor[gray]{0.9}\textbf{0.40} & \cellcolor[gray]{0.9}\textbf{0.99} &
  \cellcolor[gray]{0.9}\underline{0.56} & \cellcolor[gray]{0.9}0.32 & \cellcolor[gray]{0.9}\textbf{0.11} & \cellcolor[gray]{0.9}0.67 &
  \cellcolor[gray]{0.9}\underline{0.87} & \cellcolor[gray]{0.9}\textbf{0.72} & \cellcolor[gray]{0.9}\underline{0.22} & \cellcolor[gray]{0.9}\textbf{0.97} \\
\bottomrule
\end{tabular}
\end{adjustbox}
\caption{Ablation on surrogate-set selection. The fully gray PE\(^{+}\) line is our final patch-diverse surrogate set (\textit{CLIP$^\dag$-G/14, CLIP-B/16, CLIP-B/32, CLIP$^\dag$-B/32}). Bold numbers denote the best score; underlined numbers denote second best.}
\label{tab:ensemble_ablation}
\end{table*}

Recent ensemble attacks mainly design aggregation schemes over a \emph{small} set of surrogates~\citep{anyattack,advdiffusionvlm}. We instead start from a larger pool of CLIP, DinoV2~\citep{dinov2}, and BLIP-2~\citep{blip2} encoders (\cref{tab:surrogates}) and find that \emph{which} models are included matters: pre-selecting a few strong and complementary surrogates yields clear gains even with plain averaging. Since \emph{cross-patch-size} transfer is the hardest (\cref{tab:embedding trans}), we keep models that transfer well and span diverse patch sizes. The two-stage procedure below yields \emph{Patch Ensemble$^{+}$} (PE$^{+}$): CLIP$^\dag$-G/14, CLIP-B/16, CLIP-B/32, and CLIP$^\dag$-B/32, whose attention concentrates on the main object (\cref{fig:attention map}).

Exhaustively testing all model combinations is computationally infeasible, so we select the ensemble with a heuristic two-stage procedure guided by the embedding-transferability profile in \cref{tab:embedding trans}. We first exclude Dino-Large and BLIP-2 due to their poor transferability.

\paragraph{Stage 1: homogeneous pairs.} \Cref{tab:pair_ensemble_ablation} compares two-model groups with the same patch size. The patch-32 CLIP pair performs best on Claude 3.7, while GPT-4o and Gemini 2.5 favor patch sizes 14 and 16, showing that a fixed patch size can reduce architectural mismatch but still inherits its own patch-size bias.

\paragraph{Stage 2: patch-diverse mixtures.} Starting from the patch-32 CLIP pair, which performs strongly on Claude and transfers consistently across patch-16 and patch-32 models, we add one patch-14 and one patch-16 candidate. \Cref{tab:ensemble_ablation} evaluates these four-model mixtures. The resulting PE$^{+}$, comprising CLIP$^\dag$-G/14, CLIP-B/16, CLIP-B/32, and CLIP$^\dag$-B/32, achieves the most balanced performance across GPT-4o, Claude 3.7, and Gemini 2.5. \Cref{fig:attention map} contrasts the attention maps of selected and discarded models.

\section{Additional Ablations} \label{appendix:ablations}
\begin{figure}[tp]
    \centering
    \includegraphics[width=\linewidth]{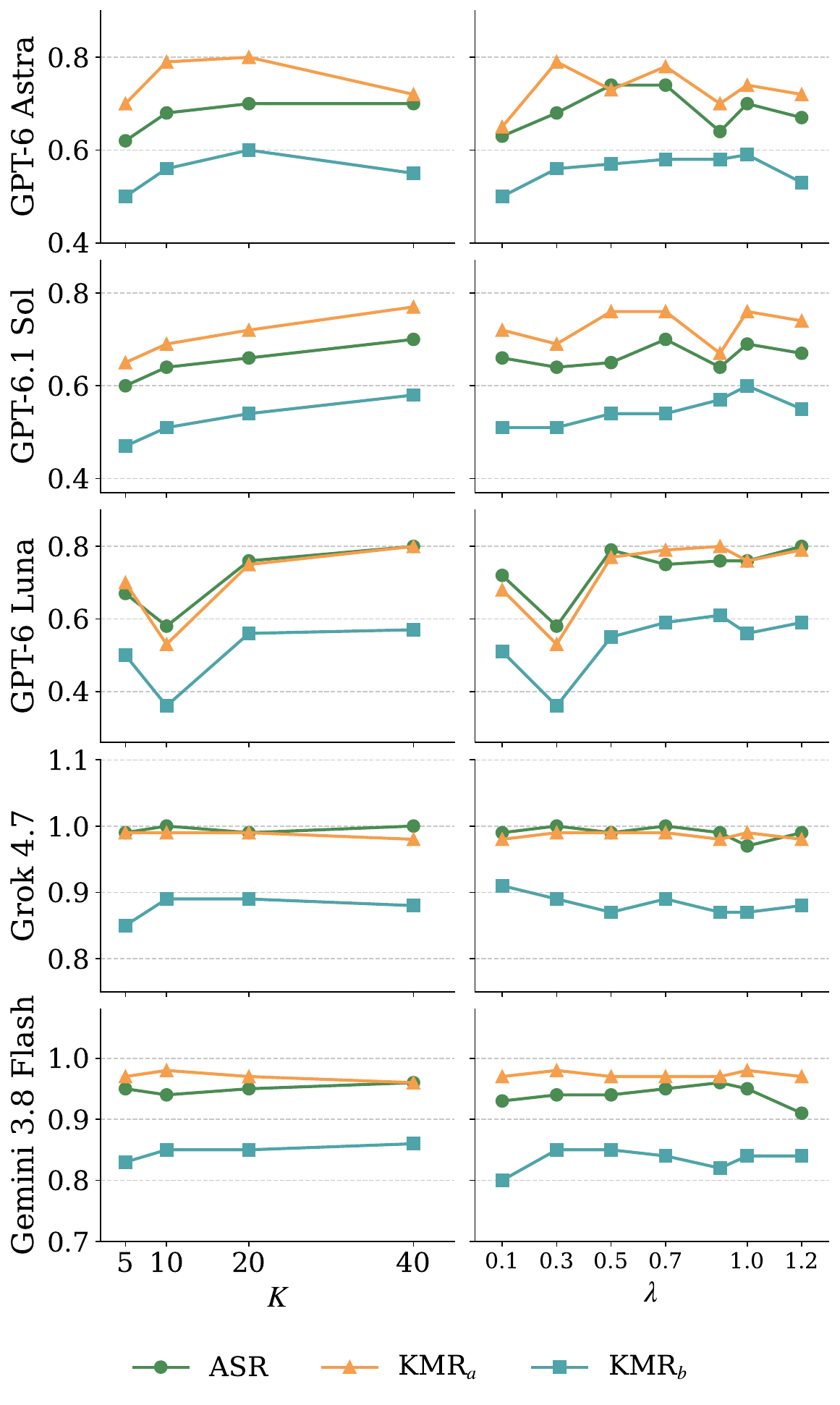}
    \caption{Hyperparameter sensitivity of \ours{}: number of crops $K$ (left) and auxiliary weight $\lambda$ (right).}
    \label{fig:k_lambda}
\end{figure}
\begin{figure*}[tp]
    \centering
    \includegraphics[width=0.8\textwidth]{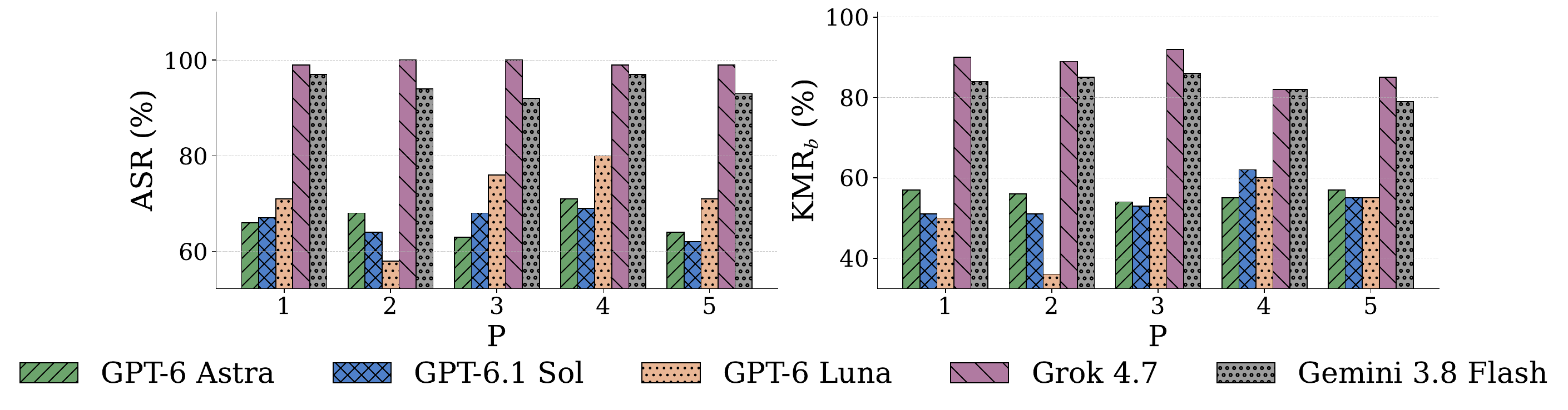}\\[0.3em]
    \includegraphics[width=0.8\textwidth]{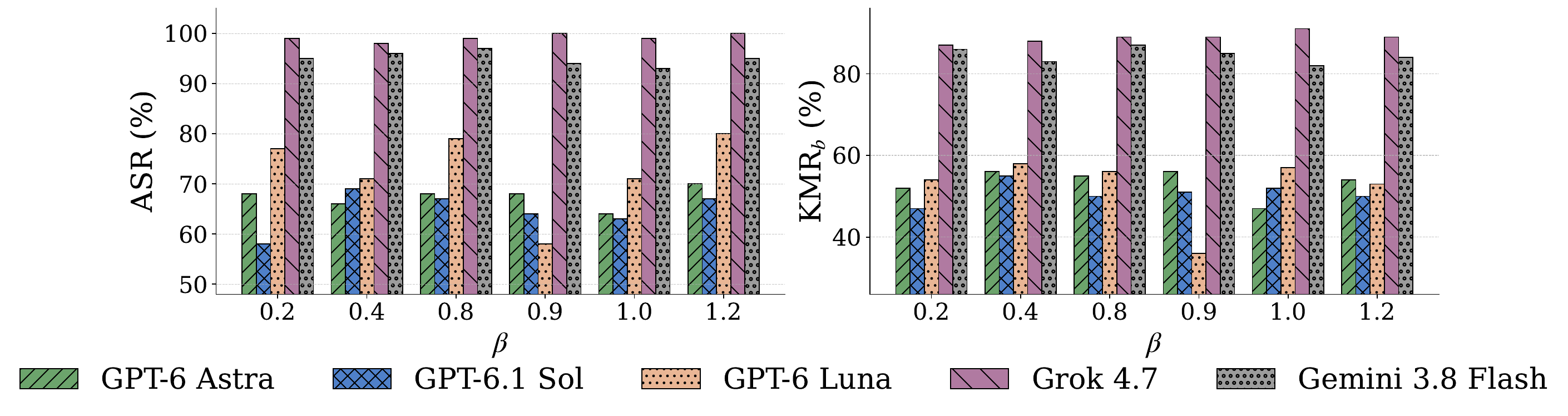}
    \caption{Hyperparameter sensitivity of \ours{}: auxiliary set size $P$ (top) and momentum parameter $\beta$ (bottom).}
    \label{fig:plot_p_gamma}
\end{figure*}
\begin{table*}[t]
\centering
\renewcommand\arraystretch{1.05}
\begin{adjustbox}{width=1.00\textwidth,center}
\scriptsize
\setlength{\tabcolsep}{1.35pt}
\begin{tabular}{p{0.95cm}<{\centering}|c|cccc|cccc|cccc}
\toprule
\multirow{2}{*}{$\alpha$} & \multirow{2}{*}{Method} &
\multicolumn{4}{c|}{\textbf{GPT-6 Astra}} &
\multicolumn{4}{c|}{\textbf{GPT-6.1 Sol}} &
\multicolumn{4}{c}{\textbf{GPT-6 Luna}} \\
\cmidrule{3-14}
& & $\text{KMR}_a$ & $\text{KMR}_b$ & $\text{KMR}_c$ & ASR & $\text{KMR}_a$ & $\text{KMR}_b$ & $\text{KMR}_c$ & ASR & $\text{KMR}_a$ & $\text{KMR}_b$ & $\text{KMR}_c$ & ASR \\
\midrule
\multirow{3}{*}{0.25}
& SSA-CWA~\citep{attackbard} & 0.22 & 0.11 & 0.01 & 0.08 & 0.17 & 0.07 & 0.01 & 0.13 & 0.26 & 0.18 & 0.05 & 0.19 \\
& \attackname{}~\citep{mattack} & 0.38 & 0.25 & 0.05 & 0.27 & 0.44 & 0.26 & 0.06 & 0.29 & 0.41 & 0.23 & 0.06 & 0.31 \\
& \ours{} (Ours) & \cellcolor[gray]{0.9}0.75 & \cellcolor[gray]{0.9}0.50 & \cellcolor[gray]{0.9}0.19 & \cellcolor[gray]{0.9}0.64 & \cellcolor[gray]{0.9}0.77 & \cellcolor[gray]{0.9}0.57 & \cellcolor[gray]{0.9}0.19 & \cellcolor[gray]{0.9}0.66 & \cellcolor[gray]{0.9}0.82 & \cellcolor[gray]{0.9}0.57 & \cellcolor[gray]{0.9}0.24 & \cellcolor[gray]{0.9}0.75 \\
\midrule
\multirow{3}{*}{0.50}
& SSA-CWA~\citep{attackbard} & 0.26 & 0.13 & 0.02 & 0.10 & 0.25 & 0.13 & 0.03 & 0.13 & 0.25 & 0.17 & 0.03 & 0.19 \\
& \attackname{}~\citep{mattack} & 0.47 & 0.34 & 0.08 & 0.36 & 0.44 & 0.25 & 0.05 & 0.32 & 0.50 & 0.39 & 0.06 & 0.41 \\
& \ours{} (Ours) & \cellcolor[gray]{0.9}0.77 & \cellcolor[gray]{0.9}0.52 & \cellcolor[gray]{0.9}0.21 & \cellcolor[gray]{0.9}0.68 & \cellcolor[gray]{0.9}0.74 & \cellcolor[gray]{0.9}0.50 & \cellcolor[gray]{0.9}0.23 & \cellcolor[gray]{0.9}0.69 & \cellcolor[gray]{0.9}0.81 & \cellcolor[gray]{0.9}0.56 & \cellcolor[gray]{0.9}0.23 & \cellcolor[gray]{0.9}0.79 \\
\midrule
\multirow{3}{*}{1.00}
& SSA-CWA~\citep{attackbard} & 0.23 & 0.11 & 0.02 & 0.11 & 0.26 & 0.09 & 0.02 & 0.12 & 0.24 & 0.09 & 0.02 & 0.13 \\
& \attackname{}~\citep{mattack} & 0.49 & 0.35 & 0.11 & 0.35 & 0.44 & 0.31 & 0.06 & 0.31 & 0.25 & 0.12 & 0.02 & 0.21 \\
& \ours{} (Ours) & \cellcolor[gray]{0.9}0.77 & \cellcolor[gray]{0.9}0.55 & \cellcolor[gray]{0.9}0.23 & \cellcolor[gray]{0.9}0.70 & \cellcolor[gray]{0.9}0.79 & \cellcolor[gray]{0.9}0.61 & \cellcolor[gray]{0.9}0.22 & \cellcolor[gray]{0.9}0.72 & \cellcolor[gray]{0.9}0.81 & \cellcolor[gray]{0.9}0.61 & \cellcolor[gray]{0.9}0.29 & \cellcolor[gray]{0.9}0.75 \\
\midrule
\multirow{3}{*}{1.275}
& SSA-CWA~\citep{attackbard} & 0.28 & 0.15 & 0.01 & 0.12 & 0.22 & 0.10 & 0.00 & 0.10 & 0.24 & 0.14 & 0.03 & 0.17 \\
& \attackname{}~\citep{mattack} & 0.54 & 0.35 & 0.09 & 0.37 & 0.43 & 0.29 & 0.11 & 0.34 & 0.43 & 0.31 & 0.09 & 0.39 \\
& \ours{} (Ours) & \cellcolor[gray]{0.9}0.78 & \cellcolor[gray]{0.9}0.59 & \cellcolor[gray]{0.9}0.28 & \cellcolor[gray]{0.9}0.69 & \cellcolor[gray]{0.9}0.72 & \cellcolor[gray]{0.9}0.52 & \cellcolor[gray]{0.9}0.18 & \cellcolor[gray]{0.9}0.69 & \cellcolor[gray]{0.9}0.66 & \cellcolor[gray]{0.9}0.51 & \cellcolor[gray]{0.9}0.19 & \cellcolor[gray]{0.9}0.72 \\
\bottomrule
\end{tabular}
\end{adjustbox}
\par\medskip
\begin{adjustbox}{width=0.71\textwidth,center}
\scriptsize
\setlength{\tabcolsep}{1.35pt}
\begin{tabular}{p{0.95cm}<{\centering}|c|cccc|cccc}
\toprule
\multirow{2}{*}{$\alpha$} & \multirow{2}{*}{Method} &
\multicolumn{4}{c|}{\textbf{Grok 4.7}} &
\multicolumn{4}{c}{\textbf{Gemini 3.8 Flash}} \\
\cmidrule{3-10}
& & $\text{KMR}_a$ & $\text{KMR}_b$ & $\text{KMR}_c$ & ASR & $\text{KMR}_a$ & $\text{KMR}_b$ & $\text{KMR}_c$ & ASR \\
\midrule
\multirow{3}{*}{0.25}
& SSA-CWA~\citep{attackbard} & 0.46 & 0.32 & 0.09 & 0.28 & 0.25 & 0.15 & 0.02 & 0.08 \\
& \attackname{}~\citep{mattack} & 0.76 & 0.51 & 0.13 & 0.67 & 0.66 & 0.40 & 0.09 & 0.46 \\
& \ours{} (Ours) & \cellcolor[gray]{0.9}0.97 & \cellcolor[gray]{0.9}0.83 & \cellcolor[gray]{0.9}0.38 & \cellcolor[gray]{0.9}0.99 & \cellcolor[gray]{0.9}0.96 & \cellcolor[gray]{0.9}0.78 & \cellcolor[gray]{0.9}0.36 & \cellcolor[gray]{0.9}0.90 \\
\midrule
\multirow{3}{*}{0.50}
& SSA-CWA~\citep{attackbard} & 0.42 & 0.29 & 0.09 & 0.30 & 0.27 & 0.16 & 0.03 & 0.12 \\
& \attackname{}~\citep{mattack} & 0.92 & 0.72 & 0.32 & 0.89 & 0.78 & 0.58 & 0.22 & 0.71 \\
& \ours{} (Ours) & \cellcolor[gray]{0.9}0.98 & \cellcolor[gray]{0.9}0.85 & \cellcolor[gray]{0.9}0.49 & \cellcolor[gray]{0.9}0.99 & \cellcolor[gray]{0.9}0.95 & \cellcolor[gray]{0.9}0.81 & \cellcolor[gray]{0.9}0.45 & \cellcolor[gray]{0.9}0.94 \\
\midrule
\multirow{3}{*}{1.00}
& SSA-CWA~\citep{attackbard} & 0.43 & 0.27 & 0.09 & 0.26 & 0.22 & 0.15 & 0.01 & 0.10 \\
& \attackname{}~\citep{mattack} & 0.96 & 0.75 & 0.32 & 0.96 & 0.92 & 0.74 & 0.30 & 0.87 \\
& \ours{} (Ours) & \cellcolor[gray]{0.9}0.99 & \cellcolor[gray]{0.9}0.89 & \cellcolor[gray]{0.9}0.51 & \cellcolor[gray]{0.9}0.98 & \cellcolor[gray]{0.9}0.97 & \cellcolor[gray]{0.9}0.79 & \cellcolor[gray]{0.9}0.42 & \cellcolor[gray]{0.9}0.95 \\
\midrule
\multirow{3}{*}{1.275}
& SSA-CWA~\citep{attackbard} & 0.39 & 0.25 & 0.05 & 0.28 & 0.26 & 0.16 & 0.01 & 0.10 \\
& \attackname{}~\citep{mattack} & 0.96 & 0.78 & 0.42 & 0.94 & 0.91 & 0.70 & 0.36 & 0.86 \\
& \ours{} (Ours) & \cellcolor[gray]{0.9}0.99 & \cellcolor[gray]{0.9}0.85 & \cellcolor[gray]{0.9}0.51 & \cellcolor[gray]{0.9}1.00 & \cellcolor[gray]{0.9}0.97 & \cellcolor[gray]{0.9}0.85 & \cellcolor[gray]{0.9}0.49 & \cellcolor[gray]{0.9}1.00 \\
\bottomrule
\end{tabular}
\end{adjustbox}
\caption{Ablation study on the impact of step size~($\alpha$).}
\label{tab:ablation-alpha}
\end{table*}
\begin{table*}[t]
\centering
\renewcommand\arraystretch{1.05}
\begin{adjustbox}{width=1.00\textwidth,center}
\scriptsize
\setlength{\tabcolsep}{1.35pt}
\begin{tabular}{p{0.95cm}<{\centering}|c|cccc|cccc|cccc}
\toprule
\multirow{2}{*}{$\epsilon$} & \multirow{2}{*}{Method} &
\multicolumn{4}{c|}{\textbf{GPT-6 Astra}} &
\multicolumn{4}{c|}{\textbf{GPT-6.1 Sol}} &
\multicolumn{4}{c}{\textbf{GPT-6 Luna}} \\
\cmidrule{3-14}
& & $\text{KMR}_a$ & $\text{KMR}_b$ & $\text{KMR}_c$ & ASR & $\text{KMR}_a$ & $\text{KMR}_b$ & $\text{KMR}_c$ & ASR & $\text{KMR}_a$ & $\text{KMR}_b$ & $\text{KMR}_c$ & ASR \\
\midrule
\multirow{5}{*}{4}
& AttackVLM~\citep{attackvlm} & 0.12 & 0.05 & 0.00 & 0.02 & 0.14 & 0.06 & 0.00 & 0.02 & 0.13 & 0.07 & 0.00 & 0.00 \\
& SSA-CWA~\citep{attackbard} & 0.14 & 0.07 & 0.01 & 0.04 & 0.14 & 0.07 & 0.01 & 0.02 & 0.19 & 0.09 & 0.01 & 0.05 \\
& AnyAttack~\citep{anyattack} & 0.19 & 0.09 & 0.00 & 0.00 & 0.19 & 0.09 & 0.00 & 0.00 & 0.19 & 0.08 & 0.00 & 0.02 \\
& \attackname{}~\citep{mattack} & 0.21 & 0.10 & 0.00 & 0.04 & 0.20 & 0.10 & 0.00 & 0.08 & 0.21 & 0.12 & 0.00 & 0.06 \\
& \ours{} (Ours) & \cellcolor[gray]{0.9}\textbf{0.32} & \cellcolor[gray]{0.9}\textbf{0.20} & \cellcolor[gray]{0.9}\textbf{0.04} & \cellcolor[gray]{0.9}\textbf{0.13} & \cellcolor[gray]{0.9}\textbf{0.37} & \cellcolor[gray]{0.9}\textbf{0.22} & \cellcolor[gray]{0.9}\textbf{0.05} & \cellcolor[gray]{0.9}\textbf{0.20} & \cellcolor[gray]{0.9}\textbf{0.32} & \cellcolor[gray]{0.9}\textbf{0.21} & \cellcolor[gray]{0.9}\textbf{0.04} & \cellcolor[gray]{0.9}\textbf{0.19} \\
\midrule
\multirow{5}{*}{8}
& AttackVLM~\citep{attackvlm} & 0.14 & 0.06 & 0.00 & 0.01 & 0.13 & 0.05 & 0.00 & 0.02 & 0.14 & 0.06 & 0.00 & 0.02 \\
& SSA-CWA~\citep{attackbard} & 0.17 & 0.05 & 0.01 & 0.02 & 0.16 & 0.11 & 0.01 & 0.07 & 0.18 & 0.07 & 0.00 & 0.05 \\
& AnyAttack~\citep{anyattack} & 0.26 & 0.12 & 0.00 & 0.03 & 0.24 & 0.10 & 0.03 & 0.05 & 0.25 & 0.13 & 0.00 & 0.07 \\
& \attackname{}~\citep{mattack} & 0.34 & 0.20 & 0.06 & 0.23 & 0.36 & 0.18 & 0.06 & 0.18 & 0.24 & 0.14 & 0.02 & 0.12 \\
& \ours{} (Ours) & \cellcolor[gray]{0.9}\textbf{0.55} & \cellcolor[gray]{0.9}\textbf{0.36} & \cellcolor[gray]{0.9}\textbf{0.13} & \cellcolor[gray]{0.9}\textbf{0.44} & \cellcolor[gray]{0.9}\textbf{0.58} & \cellcolor[gray]{0.9}\textbf{0.42} & \cellcolor[gray]{0.9}\textbf{0.14} & \cellcolor[gray]{0.9}\textbf{0.51} & \cellcolor[gray]{0.9}\textbf{0.61} & \cellcolor[gray]{0.9}\textbf{0.42} & \cellcolor[gray]{0.9}\textbf{0.12} & \cellcolor[gray]{0.9}\textbf{0.49} \\
\midrule
\multirow{5}{*}{16}
& AttackVLM~\citep{attackvlm} & 0.14 & 0.07 & 0.00 & 0.02 & 0.15 & 0.06 & 0.00 & 0.03 & 0.18 & 0.09 & 0.00 & 0.02 \\
& SSA-CWA~\citep{attackbard} & 0.23 & 0.11 & 0.02 & 0.11 & 0.26 & 0.09 & 0.02 & 0.12 & 0.24 & 0.09 & 0.02 & 0.13 \\
& AnyAttack~\citep{anyattack} & 0.48 & 0.24 & 0.05 & 0.28 & 0.49 & 0.29 & 0.05 & 0.19 & 0.32 & 0.17 & 0.02 & 0.22 \\
& \attackname{}~\citep{mattack} & 0.51 & 0.34 & 0.10 & 0.40 & 0.46 & 0.30 & 0.06 & 0.40 & 0.34 & 0.12 & 0.02 & 0.24 \\
& \ours{} (Ours) & \cellcolor[gray]{0.9}\textbf{0.79} & \cellcolor[gray]{0.9}\textbf{0.56} & \cellcolor[gray]{0.9}\textbf{0.21} & \cellcolor[gray]{0.9}\textbf{0.68} & \cellcolor[gray]{0.9}\textbf{0.69} & \cellcolor[gray]{0.9}\textbf{0.51} & \cellcolor[gray]{0.9}\textbf{0.21} & \cellcolor[gray]{0.9}\textbf{0.64} & \cellcolor[gray]{0.9}\textbf{0.53} & \cellcolor[gray]{0.9}\textbf{0.36} & \cellcolor[gray]{0.9}\textbf{0.15} & \cellcolor[gray]{0.9}\textbf{0.58} \\
\bottomrule
\end{tabular}
\end{adjustbox}
\par\medskip
\begin{adjustbox}{width=0.86\textwidth,center}
\scriptsize
\setlength{\tabcolsep}{1.35pt}
\begin{tabular}{p{0.95cm}<{\centering}|c|cccc|cccc|cc}
\toprule
\multirow{2}{*}{$\epsilon$} & \multirow{2}{*}{Method} &
\multicolumn{4}{c|}{\textbf{Grok 4.7}} &
\multicolumn{4}{c|}{\textbf{Gemini 3.8 Flash}} &
\multicolumn{2}{c}{Imperceptibility} \\
\cmidrule{3-12}
& & $\text{KMR}_a$ & $\text{KMR}_b$ & $\text{KMR}_c$ & ASR & $\text{KMR}_a$ & $\text{KMR}_b$ & $\text{KMR}_c$ & ASR & $\ell_1\!\downarrow$ & $\ell_2\!\downarrow$ \\
\midrule
\multirow{5}{*}{4}
& AttackVLM~\citep{attackvlm} & 0.13 & 0.06 & 0.00 & 0.00 & 0.14 & 0.05 & 0.00 & 0.00 & 0.011 & 0.012 \\
& SSA-CWA~\citep{attackbard} & 0.12 & 0.06 & 0.00 & 0.01 & 0.14 & 0.06 & 0.00 & 0.01 & 0.015 & 0.016 \\
& AnyAttack~\citep{anyattack} & 0.21 & 0.09 & 0.00 & 0.01 & 0.15 & 0.05 & 0.00 & 0.01 & 0.015 & 0.015 \\
& \attackname{}~\citep{mattack} & 0.33 & 0.19 & 0.04 & 0.24 & 0.24 & 0.13 & 0.02 & 0.09 & \textbf{0.009} & \textbf{0.011} \\
& \ours{} (Ours) & \cellcolor[gray]{0.9}\textbf{0.68} & \cellcolor[gray]{0.9}\textbf{0.52} & \cellcolor[gray]{0.9}\textbf{0.15} & \cellcolor[gray]{0.9}\textbf{0.57} & \cellcolor[gray]{0.9}\textbf{0.47} & \cellcolor[gray]{0.9}\textbf{0.26} & \cellcolor[gray]{0.9}\textbf{0.05} & \cellcolor[gray]{0.9}\textbf{0.20} & \cellcolor[gray]{0.9}0.013 & \cellcolor[gray]{0.9}0.014 \\
\midrule
\multirow{5}{*}{8}
& AttackVLM~\citep{attackvlm} & 0.13 & 0.06 & 0.00 & 0.00 & 0.14 & 0.05 & 0.00 & 0.01 & 0.021 & 0.023 \\
& SSA-CWA~\citep{attackbard} & 0.22 & 0.08 & 0.00 & 0.07 & 0.16 & 0.07 & 0.02 & 0.02 & 0.030 & 0.031 \\
& AnyAttack~\citep{anyattack} & 0.34 & 0.18 & 0.03 & 0.09 & 0.34 & 0.13 & 0.01 & 0.09 & 0.029 & 0.030 \\
& \attackname{}~\citep{mattack} & 0.83 & 0.60 & 0.20 & 0.75 & 0.67 & 0.45 & 0.13 & 0.47 & \textbf{0.018} & \textbf{0.020} \\
& \ours{} (Ours) & \cellcolor[gray]{0.9}\textbf{0.95} & \cellcolor[gray]{0.9}\textbf{0.82} & \cellcolor[gray]{0.9}\textbf{0.40} & \cellcolor[gray]{0.9}\textbf{0.98} & \cellcolor[gray]{0.9}\textbf{0.84} & \cellcolor[gray]{0.9}\textbf{0.64} & \cellcolor[gray]{0.9}\textbf{0.28} & \cellcolor[gray]{0.9}\textbf{0.69} & \cellcolor[gray]{0.9}0.023 & \cellcolor[gray]{0.9}0.025 \\
\midrule
\multirow{5}{*}{16}
& AttackVLM~\citep{attackvlm} & 0.13 & 0.04 & 0.00 & 0.01 & 0.15 & 0.06 & 0.01 & 0.01 & 0.037 & 0.042 \\
& SSA-CWA~\citep{attackbard} & 0.43 & 0.27 & 0.09 & 0.26 & 0.22 & 0.15 & 0.01 & 0.10 & 0.060 & 0.061 \\
& AnyAttack~\citep{anyattack} & 0.67 & 0.41 & 0.12 & 0.49 & 0.63 & 0.40 & 0.06 & 0.34 & 0.054 & 0.057 \\
& \attackname{}~\citep{mattack} & 0.97 & 0.77 & 0.40 & 0.95 & 0.95 & 0.74 & 0.33 & 0.88 & \textbf{0.031} & \textbf{0.037} \\
& \ours{} (Ours) & \cellcolor[gray]{0.9}\textbf{0.99} & \cellcolor[gray]{0.9}\textbf{0.89} & \cellcolor[gray]{0.9}\textbf{0.50} & \cellcolor[gray]{0.9}\textbf{1.00} & \cellcolor[gray]{0.9}\textbf{0.98} & \cellcolor[gray]{0.9}\textbf{0.85} & \cellcolor[gray]{0.9}\textbf{0.47} & \cellcolor[gray]{0.9}\textbf{0.94} & \cellcolor[gray]{0.9}0.039 & \cellcolor[gray]{0.9}0.044 \\
\bottomrule
\end{tabular}
\end{adjustbox}
\caption{Ablation study on the impact of perturbation budget~($\epsilon$).}
\label{tab:ablation-e}
\end{table*}
\subsection{Step-Size Sensitivity}
\Cref{tab:ablation-alpha} reports the full sweep over step size $\alpha$. \ours{} is the strongest method at every step size on all five victims, and its ASR varies little with $\alpha$: 0.64--0.79 on the GPT-6 models and 0.90--1.00 on Grok 4.7 and Gemini 3.8 Flash. Small steps under-explore the $\ell_\infty$ ball for \attackname{}: at $\alpha=0.25$ its ASR on Gemini 3.8 Flash is 0.46, against 0.86--0.87 at $\alpha\geq1$. The gain of \ours{} is therefore not caused by taking larger steps; MCA, ATA, and PM give consistent update directions even with small steps. We use $\alpha=1.275$ in the main setting, where \ours{} attains the highest ASR on Grok 4.7 and Gemini 3.8 Flash.

\subsection{Perturbation Budget}
\Cref{tab:ablation-e} evaluates $\epsilon\in\{4,8,16\}$ and reports both attack metrics and perturbation norms. \ours{} is the strongest method on every metric for all five victims at each budget. The gap is largest at $\epsilon=8$, where \ours{} reaches 44--51\% ASR on the GPT-6 models against at most 23\% for \attackname{}, and 98\% on Grok 4.7. At $\epsilon=4$ every attack stays weak on the GPT-6 models (at most 20\% ASR). The imperceptibility columns and the user study in \cref{subsec:human_perceptual} further indicate that the improved transferability is achieved with perturbations that remain hard to detect.

\subsection{Hyperparameter Sensitivity} \label{appendix:other params}
\Cref{fig:trajectory} shows that MCA accelerates convergence: $K=10$ crops already improve the optimization trajectory over single-crop alignment ($K=1$), with additional margin from $K=100$. \Cref{fig:k_lambda} sweeps $K$ and $\lambda$ on the five victims. Grok 4.7 and Gemini 3.8 Flash are saturated, with ASR between 0.91 and 1.00 at every setting. On GPT-6 Astra and GPT-6.1 Sol, ASR grows with the number of crops, from 0.62 and 0.60 at $K=5$ to 0.70 at $K=40$, and $\lambda$ between 0.3 and 1.0 changes ASR by at most 0.10. GPT-6 Luna varies the most across settings (0.58--0.80 ASR) without a consistent trend. \Cref{fig:plot_p_gamma} shows smaller effects of $P$ and the momentum coefficient $\beta$: on GPT-6 Astra and GPT-6.1 Sol, ASR stays within 0.58--0.71 for $P\in\{1,\dots,5\}$ and $\beta\in[0.2,1.2]$. We keep $P=2$ for efficiency and $\beta=0.9$ as in the main experiments.

\subsection{Step Budget on Other Victims} \label{appendix:steps}
\begin{figure}[tp]
    \centering
    \includegraphics[width=\linewidth,trim=0 20 0 0,clip]{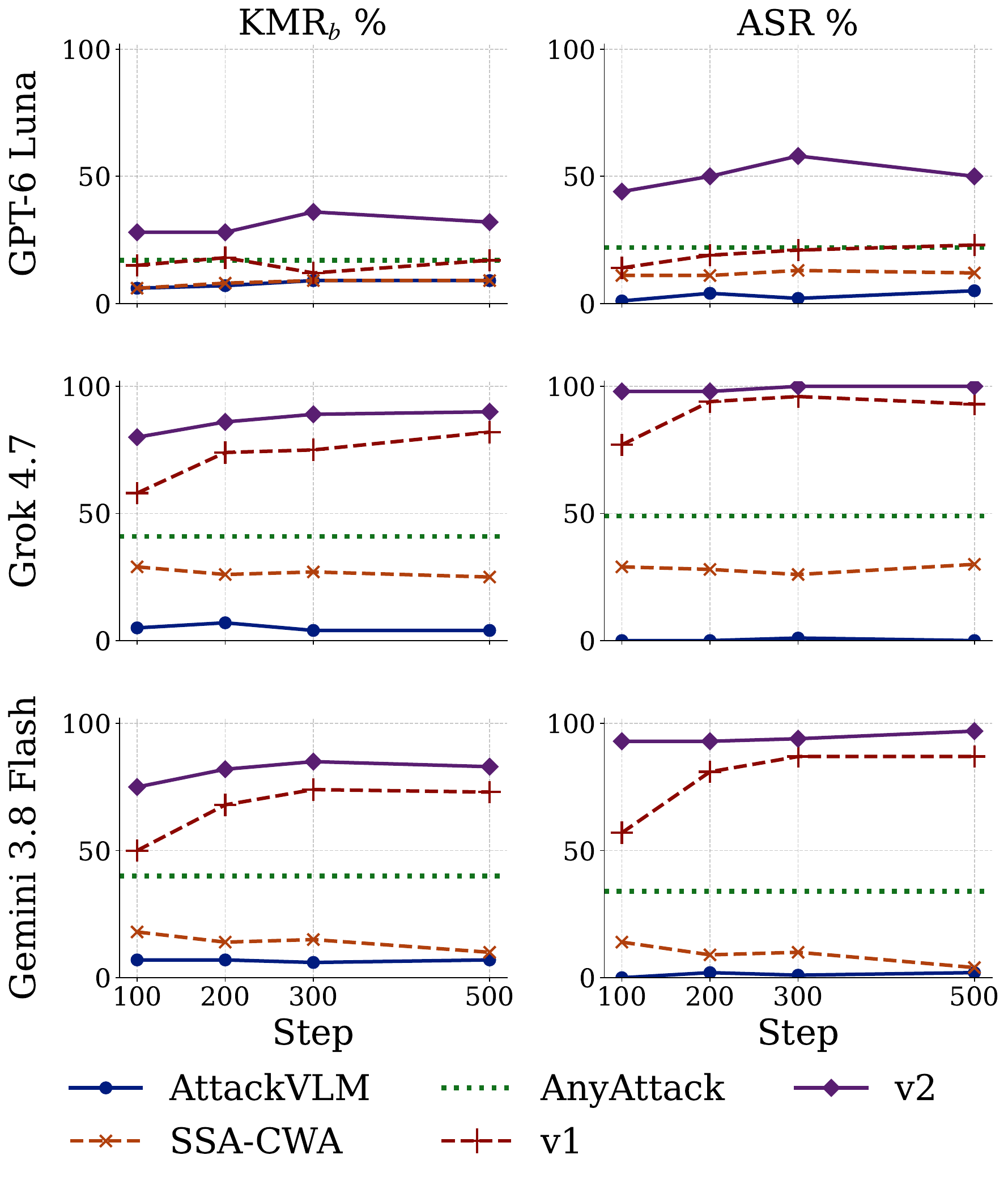}
    \caption{Comparison under different step budgets on GPT-6 Luna, Grok 4.7, and Gemini 3.8 Flash.}
    \label{fig:step_appendix}
\end{figure}

\Cref{fig:step_appendix} repeats the step-budget comparison of \cref{fig:ablation_step} on GPT-6 Luna, Grok 4.7, and Gemini 3.8 Flash. \ours{} stays above \attackname{} at every budget on all three victims, and with 100 steps its ASR already exceeds that of \attackname{} at 500 steps.

\section{Additional Results} \label{appendix:results}
\begin{table*}[tp]
\centering
\renewcommand\arraystretch{1.1}
\begin{adjustbox}{width=\textwidth,center}
\small
\setlength{\tabcolsep}{2.5pt}
\begin{tabular}{c|c|cccc|cccc|cccc}
\toprule
\multirow{2}{*}{\textbf{Method}} & \multirow{2}{*}{\textbf{Model}} &
\multicolumn{4}{c|}{\textbf{GPT-6 Luna}} &
\multicolumn{4}{c|}{\textbf{Grok 4.7}} &
\multicolumn{4}{c}{\textbf{Gemini 3.8 Flash}} \\
\cmidrule{3-14}
& & $\text{KMR}_a$ & $\text{KMR}_b$ & $\text{KMR}_c$ & ASR
  & $\text{KMR}_a$ & $\text{KMR}_b$ & $\text{KMR}_c$ & ASR
  & $\text{KMR}_a$ & $\text{KMR}_b$ & $\text{KMR}_c$ & ASR \\
\midrule
\multirow{3}{*}{AttackVLM~\citep{attackvlm}} & \textbf{B/16} & 0.11 & 0.05 & 0.00 & 0.04 & 0.16 & 0.07 & 0.01 & 0.05 & 0.15 & 0.07 & 0.00 & 0.02 \\
 & \textbf{B/32} & 0.18 & 0.09 & 0.00 & 0.02 & 0.13 & 0.04 & 0.00 & 0.01 & 0.15 & 0.06 & 0.01 & 0.01 \\
 & \textbf{Laion}$^\dagger$ & 0.14 & 0.04 & 0.00 & 0.04 & 0.13 & 0.07 & 0.00 & 0.01 & 0.15 & 0.08 & 0.01 & 0.02 \\
\midrule
SSA-CWA~\citep{attackbard} & \textbf{Ensemble} & 0.24 & 0.09 & 0.02 & 0.13 & 0.43 & 0.27 & 0.09 & 0.26 & 0.22 & 0.15 & 0.01 & 0.10 \\
AnyAttack~\citep{anyattack} & \textbf{Ensemble} & 0.32 & 0.17 & 0.02 & 0.22 & 0.67 & 0.41 & 0.12 & 0.49 & 0.63 & 0.40 & 0.06 & 0.34 \\
FOA-Attack~\citep{jia2025adversarial} & \textbf{Ensemble} & 0.33 & 0.21 & 0.00 & 0.22 & 0.96 & 0.82 & 0.41 & 0.97 & 0.93 & 0.77 & 0.34 & 0.92 \\
SGHA-Attack~\citep{wang2026sgha} & \textbf{Ensemble} & 0.14 & 0.04 & 0.00 & 0.04 & 0.17 & 0.08 & 0.01 & 0.02 & 0.14 & 0.07 & 0.01 & 0.03 \\
\textbf{\attackname{}}~\citep{mattack} & \textbf{Ensemble} & 0.34 & 0.12 & 0.02 & 0.24 & 0.97 & 0.77 & 0.40 & 0.95 & 0.95 & 0.74 & 0.33 & 0.88 \\
\textbf{\ours{}} (Ours) & \textbf{Ensemble} & \cellcolor[gray]{0.9}0.78 & \cellcolor[gray]{0.9}0.57 & \cellcolor[gray]{0.9}0.18 & \cellcolor[gray]{0.9}0.76 & \cellcolor[gray]{0.9}0.95 & \cellcolor[gray]{0.9}0.83 & \cellcolor[gray]{0.9}0.47 & \cellcolor[gray]{0.9}0.98 & \cellcolor[gray]{0.9}0.96 & \cellcolor[gray]{0.9}0.79 & \cellcolor[gray]{0.9}0.35 & \cellcolor[gray]{0.9}0.95 \\
\midrule
MPCAttack~\citep{li2026multi} & \textbf{Ens.+LVLM} & 0.50 & 0.37 & 0.15 & 0.49 & \textbf{0.99} & \textbf{0.88} & 0.50 & \textbf{1.00} & \textbf{0.99} & \textbf{0.93} & \textbf{0.55} & \textbf{1.00} \\
\textbf{\ours{}} (Ours) & \textbf{Ens.+LVLM} & \cellcolor[gray]{0.9}\textbf{0.81} & \cellcolor[gray]{0.9}\textbf{0.61} & \cellcolor[gray]{0.9}\textbf{0.24} & \cellcolor[gray]{0.9}\textbf{0.83} & \cellcolor[gray]{0.9}0.97 & \cellcolor[gray]{0.9}0.87 & \cellcolor[gray]{0.9}\textbf{0.54} & \cellcolor[gray]{0.9}0.99 & \cellcolor[gray]{0.9}0.98 & \cellcolor[gray]{0.9}0.89 & \cellcolor[gray]{0.9}0.49 & \cellcolor[gray]{0.9}0.99 \\
\bottomrule
\end{tabular}
\end{adjustbox}
\caption{Comparison of attack methods on GPT-6 Luna, Grok 4.7, and Gemini 3.8 Flash, complementing \cref{tab:main-compare}. $^\dagger$: pre-trained on LAION~\citep{laion5b}. Ens.+LVLM: CLIP ensemble plus InternVL3-1B and DINOv2.}
\label{tab:grok-gemini}
\end{table*}
\begin{table}[t]
\centering
\begin{adjustbox}{width=\columnwidth,center}
\begin{tabular}{l|l|cccccc}
\toprule
\textbf{Method} & \textbf{Surrogates} & \textbf{Astra} & \textbf{Sol} & \textbf{Claude} & \textbf{Luna} & \textbf{Grok} & \textbf{Gemini} \\
\midrule
MPCAttack & MPCAttack & 0.48 & 0.38 & 0.98 & 0.49 & \textbf{1.00} & \textbf{1.00} \\
\textbf{\ours{}} & PE$^+$ & \cellcolor[gray]{0.9}0.60 & \cellcolor[gray]{0.9}0.61 & \cellcolor[gray]{0.9}0.97 & \cellcolor[gray]{0.9}0.76 & \cellcolor[gray]{0.9}0.98 & \cellcolor[gray]{0.9}0.95 \\
\textbf{\ours{}} & Ens.+LVLM & \cellcolor[gray]{0.9}0.73 & \cellcolor[gray]{0.9}\textbf{0.75} & \cellcolor[gray]{0.9}\textbf{0.99} & \cellcolor[gray]{0.9}\textbf{0.83} & \cellcolor[gray]{0.9}0.99 & \cellcolor[gray]{0.9}0.99 \\
\textbf{\ours{}} & MPCAttack & \cellcolor[gray]{0.9}\textbf{0.74} & \cellcolor[gray]{0.9}0.62 & \cellcolor[gray]{0.9}0.98 & \cellcolor[gray]{0.9}0.81 & \cellcolor[gray]{0.9}\textbf{1.00} & \cellcolor[gray]{0.9}0.99 \\
\bottomrule
\end{tabular}
\end{adjustbox}
\caption{ASR of \ours{} with three surrogate sets, and of MPCAttack with its own, on GPT-6 Astra, GPT-6.1 Sol, Claude Sonnet 5.5, GPT-6 Luna, Grok 4.7, and Gemini 3.8 Flash ($\epsilon=16$, 300 steps). MPCAttack: CLIP-B/16, CLIP-B/32, CLIP$^\dag$-G/14, InternVL3-1B, and DINOv2-B. Ens.+LVLM: PE$^+$ plus InternVL3-1B and DINOv2-B.}
\label{tab:surrogate-sets}
\end{table}
\begin{table*}[tp]
\centering
\renewcommand\arraystretch{1.16}
\begin{adjustbox}{width=0.9\textwidth,center}
\small
\setlength{\tabcolsep}{2pt}
\begin{tabular}{c|c|cccc|cccc|cccc}
\toprule
\multirow{2}{*}{\textbf{Method}} & \multirow{2}{*}{\textbf{Model}} &
\multicolumn{4}{c|}{\textbf{GPT-5 (low)}} &
\multicolumn{4}{c|}{\textbf{GPT-5 (medium)}} &
\multicolumn{4}{c}{\textbf{GPT-5 (high)}} \\
\cmidrule{3-14}
& & $\text{KMR}_a$ & $\text{KMR}_b$ & $\text{KMR}_c$ & ASR
  & $\text{KMR}_a$ & $\text{KMR}_b$ & $\text{KMR}_c$ & ASR
  & $\text{KMR}_a$ & $\text{KMR}_b$ & $\text{KMR}_c$ & ASR \\
\midrule
SSA\mbox{-}CWA~\citep{attackbard} & \textbf{Ensemble} &
  0.08 & 0.04 & 0.00 & 0.08 &
  0.09 & 0.05 & 0.01 & 0.06 &
  0.10 & 0.05 & 0.01 & 0.07 \\
FOA\mbox{-}Attack~\citep{jia2025adversarial} & \textbf{Ensemble} &
  0.90 & 0.67 & 0.23 & 0.94 &
  0.90 & 0.69 & 0.21 & \textbf{0.96} &
  0.87 & 0.68 & 0.24 & 0.96 \\
\attackname{}~\citep{mattack} & \textbf{Ensemble} &
  0.89 & 0.65 & 0.25 & 0.98 &
  0.85 & 0.61 & 0.16 & \textbf{0.96} &
  0.80 & 0.60 & 0.20 & 0.93 \\
\textbf{\ours{}} (Ours) & \textbf{Ensemble} &
  \cellcolor[gray]{0.9}\textbf{0.92} & \cellcolor[gray]{0.9}\textbf{0.79} & \cellcolor[gray]{0.9}\textbf{0.30} & \cellcolor[gray]{0.9}\textbf{1.00} &
  \cellcolor[gray]{0.9}\textbf{0.90} & \cellcolor[gray]{0.9}\textbf{0.73} & \cellcolor[gray]{0.9}\textbf{0.25} & \cellcolor[gray]{0.9}\textbf{0.96} &
  \cellcolor[gray]{0.9}\textbf{0.88} & \cellcolor[gray]{0.9}\textbf{0.71} & \cellcolor[gray]{0.9}\textbf{0.27} & \cellcolor[gray]{0.9}\textbf{0.99} \\
\bottomrule
\end{tabular}
\end{adjustbox}
\caption{Comparison on GPT\mbox{-}5 under three reasoning-effort settings (low/medium/high).}
\label{tab:gpt5-compare-budgets}
\end{table*}
\begin{table*}[tp]
\centering
\renewcommand\arraystretch{1.1}
\begin{adjustbox}{width=\textwidth,center}
\small
\setlength{\tabcolsep}{2.5pt}
\begin{tabular}{c|c|cccc|cccc|cccc}
\toprule
\multirow{2}{*}{\textbf{Method}} & \multirow{2}{*}{\textbf{Model}} &
\multicolumn{4}{c|}{\textbf{GPT-6 Astra}} &
\multicolumn{4}{c|}{\textbf{GPT-6.1 Sol}} &
\multicolumn{4}{c}{\textbf{GPT-6 Luna}} \\
\cmidrule{3-14}
& & $\text{KMR}_a$ & $\text{KMR}_b$ & $\text{KMR}_c$ & ASR & $\text{KMR}_a$ & $\text{KMR}_b$ & $\text{KMR}_c$ & ASR & $\text{KMR}_a$ & $\text{KMR}_b$ & $\text{KMR}_c$ & ASR \\
\midrule
\multirow{3}{*}{AttackVLM~\citep{attackvlm}} & \textbf{B/16} & 0.09 & 0.04 & 0.00 & 0.02 & 0.09 & 0.04 & 0.01 & 0.03 & 0.11 & 0.05 & 0.00 & 0.03 \\
 & \textbf{B/32} & 0.10 & 0.04 & 0.00 & 0.02 & 0.10 & 0.04 & 0.01 & 0.02 & 0.10 & 0.05 & 0.01 & 0.03 \\
 & \textbf{Laion}$^\dagger$ & 0.09 & 0.04 & 0.00 & 0.02 & 0.10 & 0.04 & 0.00 & 0.02 & 0.10 & 0.05 & 0.01 & 0.03 \\
\midrule
AnyAttack~\citep{anyattack} & \textbf{Ensemble} & 0.41 & 0.25 & 0.06 & 0.22 & 0.41 & 0.26 & 0.07 & 0.23 & 0.43 & 0.27 & 0.08 & 0.29 \\
FOA-Attack~\citep{jia2025adversarial} & \textbf{Ensemble} & 0.46 & 0.31 & 0.10 & 0.36 & 0.42 & 0.28 & 0.10 & 0.33 & 0.31 & 0.17 & 0.04 & 0.24 \\
\textbf{\attackname{}}~\citep{mattack} & \textbf{Ensemble} & 0.40 & 0.26 & 0.07 & 0.28 & 0.37 & 0.23 & 0.07 & 0.26 & 0.41 & 0.27 & 0.08 & 0.32 \\
\textbf{\ours{}} (Ours) & \textbf{Ensemble} & \cellcolor[gray]{0.9}\textbf{0.62} & \cellcolor[gray]{0.9}\textbf{0.47} & \cellcolor[gray]{0.9}\textbf{0.18} & \cellcolor[gray]{0.9}\textbf{0.53} & \cellcolor[gray]{0.9}\textbf{0.60} & \cellcolor[gray]{0.9}\textbf{0.43} & \cellcolor[gray]{0.9}\textbf{0.17} & \cellcolor[gray]{0.9}\textbf{0.52} & \cellcolor[gray]{0.9}\textbf{0.63} & \cellcolor[gray]{0.9}\textbf{0.47} & \cellcolor[gray]{0.9}\textbf{0.17} & \cellcolor[gray]{0.9}\textbf{0.61} \\
\bottomrule
\end{tabular}
\end{adjustbox}
\par\medskip
\begin{adjustbox}{width=0.8\textwidth,center}
\small
\setlength{\tabcolsep}{2.5pt}
\begin{tabular}{c|c|cccc|cccc}
\toprule
\multirow{2}{*}{\textbf{Method}} & \multirow{2}{*}{\textbf{Model}} &
\multicolumn{4}{c|}{\textbf{Grok 4.7}} &
\multicolumn{4}{c}{\textbf{Gemini 3.8 Flash}} \\
\cmidrule{3-10}
& & $\text{KMR}_a$ & $\text{KMR}_b$ & $\text{KMR}_c$ & ASR & $\text{KMR}_a$ & $\text{KMR}_b$ & $\text{KMR}_c$ & ASR \\
\midrule
\multirow{3}{*}{AttackVLM~\citep{attackvlm}} & \textbf{B/16} & 0.11 & 0.06 & 0.01 & 0.04 & 0.09 & 0.04 & 0.01 & 0.02 \\
 & \textbf{B/32} & 0.09 & 0.04 & 0.01 & 0.02 & 0.09 & 0.04 & 0.01 & 0.01 \\
 & \textbf{Laion}$^\dagger$ & 0.09 & 0.04 & 0.01 & 0.03 & 0.09 & 0.04 & 0.01 & 0.01 \\
\midrule
AnyAttack~\citep{anyattack} & \textbf{Ensemble} & 0.55 & 0.37 & 0.12 & 0.39 & 0.52 & 0.33 & 0.09 & 0.27 \\
FOA-Attack~\citep{jia2025adversarial} & \textbf{Ensemble} & 0.92 & 0.79 & 0.42 & 0.91 & 0.83 & 0.68 & 0.32 & 0.78 \\
\textbf{\attackname{}}~\citep{mattack} & \textbf{Ensemble} & 0.89 & 0.74 & 0.38 & 0.87 & 0.79 & 0.65 & 0.28 & 0.72 \\
\textbf{\ours{}} (Ours) & \textbf{Ensemble} & \cellcolor[gray]{0.9}\textbf{0.96} & \cellcolor[gray]{0.9}\textbf{0.86} & \cellcolor[gray]{0.9}\textbf{0.51} & \cellcolor[gray]{0.9}\textbf{0.97} & \cellcolor[gray]{0.9}\textbf{0.89} & \cellcolor[gray]{0.9}\textbf{0.76} & \cellcolor[gray]{0.9}\textbf{0.39} & \cellcolor[gray]{0.9}\textbf{0.88} \\
\bottomrule
\end{tabular}
\end{adjustbox}
\caption{Comparison of attack methods on 1,000 source--target pairs ($\epsilon=16$, 300 steps), extending \cref{tab:main-compare,tab:grok-gemini}. KMR uses the human keywords for the 100 main targets and GPT-6 Astra keywords for the other 900 (\cref{appendix:1k}). $^\dagger$: pre-trained on LAION~\citep{laion5b}.}
\label{tab:main_table_1000}
\end{table*}

\subsection{GPT-6 Luna, Grok 4.7, and Gemini 3.8 Flash} \label{appendix:grok-gemini}
\Cref{tab:grok-gemini} extends \cref{tab:main-compare} to GPT-6 Luna, Grok 4.7, and Gemini 3.8 Flash. On GPT-6 Luna, \ours{} raises ASR from 24\% (\attackname{}) to 76\%, against 22\% for FOA-Attack. Grok 4.7 and Gemini 3.8 Flash are easier targets: FOA-Attack, \attackname{}, and \ours{} all exceed 85\% ASR there, and among the image-encoder ensembles \ours{} attains the highest ASR on both (98\% and 95\%) and the highest KMR$_b$ and KMR$_c$. With Ens.+LVLM surrogates, MPCAttack and \ours{} reach 100\% and 99\% ASR on both, while on GPT-6 Luna \ours{} reaches 83\% against 49\%.

\subsection{Effect of the Surrogate Set} \label{appendix:surrogate-sets}
\Cref{tab:surrogate-sets} runs \ours{} with three surrogate sets and every other setting fixed: PE$^+$, PE$^+$ plus InternVL3-1B and DINOv2-B (Ens.+LVLM), and the set of MPCAttack~\citep{li2026multi} (CLIP-B/16, CLIP-B/32, CLIP$^\dag$-G/14, InternVL3-1B, and DINOv2-B). The two non-CLIP models raise ASR on the GPT-6 models by 7--14 points. On the set of MPCAttack, \ours{} exceeds MPCAttack by 24--32 points on the GPT-6 models and is within one point of it on Claude Sonnet 5.5, Grok 4.7, and Gemini 3.8 Flash, where both reach at least 98\%.

\subsection{Other GPT-5 Reasoning Modes} \label{appendix:gpt5}

\Cref{tab:gpt5-compare-budgets} reports the low/medium/high GPT-5 reasoning-effort settings. \ours{} remains the strongest variant across reasoning budgets, although the gains are not strictly monotonic.

\subsection{Results on 1K Images} \label{appendix:1k}
\Cref{tab:main_table_1000} repeats the comparison of \cref{tab:main-compare,tab:grok-gemini} for the five API victims other than Claude Sonnet 5.5 on 1,000 source--target pairs from the same NIPS 2017 and target collections, with $\epsilon=16$ and 300 steps for every method. Only the 100 main targets have human-annotated keywords; GPT-6 Astra annotates the other 900 with a prompt that follows the human style (1--4 everyday words for the main subject, the action, and a prominent setting). On the 100 human-annotated targets, KMR computed with GPT-6 Astra keywords agrees with KMR from the human keywords on 92\%, 87\%, and 93\% of the per-image KMR$_a$, KMR$_b$, and KMR$_c$ decisions over the eight methods and these five victims, and the aggregate rates differ by at most 0.09. FOA-Attack uses the adversarial images released by its authors for these pairs. \ours{} attains the highest KMR and ASR on all five victims, with ASR of 53\%, 52\%, and 61\% on GPT-6 Astra, GPT-6.1 Sol, and GPT-6 Luna against at most 36\% for prior attacks, so the 100-pair ranking holds at ten times the sample size.

\subsection{Cross-Domain Evaluation} \label{sec:cross_domain_evaluation}
\begin{table*}[tp]
\centering
\renewcommand\arraystretch{1.0}
\begin{adjustbox}{width=0.80\textwidth,center}
\footnotesize
\setlength{\tabcolsep}{1.35pt}
\begin{tabular}{l|cccc|cccc|cccc}
\toprule
\multirow{2}{*}{Method} &
\multicolumn{4}{c|}{\textbf{GPT-4o}} &
\multicolumn{4}{c|}{\textbf{Claude 3.7}} &
\multicolumn{4}{c}{\textbf{Gemini 2.5}} \\
\cmidrule(lr){2-5}\cmidrule(lr){6-9}\cmidrule(lr){10-13}
& KMR$_a$ & KMR$_b$ & KMR$_c$ & ASR
& KMR$_a$ & KMR$_b$ & KMR$_c$ & ASR
& KMR$_a$ & KMR$_b$ & KMR$_c$ & ASR \\
\midrule
\multicolumn{13}{l}{\textit{PatternNet}}\\
AttackVLM & 0.06 & 0.01 & 0.00 & 0.02 & 0.06 & 0.02 & 0.00 & 0.00 & 0.09 & 0.04 & 0.00 & 0.03 \\
SSA-CWA & 0.05 & 0.02 & 0.00 & 0.13 & 0.04 & 0.03 & 0.00 & 0.07 & 0.08 & 0.02 & 0.01 & 0.15 \\
AnyAttack & 0.06 & 0.03 & 0.00 & 0.05 & 0.03 & 0.01 & 0.00 & 0.05 & 0.06 & 0.02 & 0.00 & 0.05 \\
\attackname{} & \underline{0.79} & \underline{0.66} & \underline{0.21} & \textbf{0.93} & \underline{0.33} & \underline{0.17} & \underline{0.04} & \underline{0.48} & \underline{0.86} & \textbf{0.71} & \textbf{0.23} & \underline{0.91} \\
\rowcolor[gray]{0.9}
\ours{} & \textbf{0.83} & \textbf{0.71} & \textbf{0.24} & \textbf{0.93} & \textbf{0.58} & \textbf{0.40} & \textbf{0.09} & \textbf{0.73} & \textbf{0.88} & \underline{0.68} & \underline{0.22} & \textbf{0.97} \\
\midrule
\multicolumn{13}{l}{\textit{ChestMNIST}}\\
AttackVLM & 0.06 & 0.01 & 0.00 & 0.03 & 0.05 & 0.02 & 0.00 & 0.02 & 0.08 & 0.03 & 0.00 & 0.02 \\
SSA-CWA & 0.06 & 0.03 & 0.00 & 0.15 & 0.04 & 0.03 & 0.00 & 0.07 & 0.08 & 0.02 & 0.01 & 0.14 \\
AnyAttack & 0.06 & 0.02 & 0.00 & 0.05 & 0.03 & 0.01 & 0.00 & 0.04 & 0.07 & 0.02 & 0.00 & 0.05 \\
\attackname{} & \underline{0.89} & \underline{0.70} & \underline{0.22} & \underline{0.92} & \underline{0.31} & \underline{0.18} & \underline{0.07} & \underline{0.31} & \underline{0.85} & \underline{0.67} & \underline{0.23} & \textbf{0.96} \\
\rowcolor[gray]{0.9}
\ours{} & \textbf{0.90} & \textbf{0.74} & \textbf{0.27} & \textbf{0.97} & \textbf{0.70} & \textbf{0.51} & \textbf{0.21} & \textbf{0.83} & \textbf{0.89} & \textbf{0.76} & \textbf{0.33} & \underline{0.95} \\
\bottomrule
\end{tabular}
\end{adjustbox}
\caption{Cross-domain results on \textit{PatternNet}~\citep{zhou2018patternnet} and \textit{ChestMNIST} from MedMNIST~\citep{medmnistv1}. We report KMR$_{a/b/c}$ and ASR (higher is better). \textbf{Bold} = best in column; \underline{underline} = second best. Shaded rows are our method.}
\label{tab:patternnet_cross_domain}\label{tab:chestmnist_cross_domain}
\end{table*}

We augment the \emph{NIPS 2017 adversarial competition} evaluation with medical images from \textit{ChestMNIST}~\citep{medmnistv1} and overhead imagery from \textit{PatternNet}~\citep{zhou2018patternnet}, while keeping the target set, attack budget, and optimization hyperparameters unchanged.
These domains stress a different failure mode from ordinary natural-image transfer. Medical and overhead images contain fewer object-centric cues, different texture statistics, and weaker overlap with common web-scale visual priors, so a successful targeted attack must rely less on dataset familiarity and more on stable cross-view semantic alignment. This makes the setting a useful robustness probe for our MCA/ATA/PM design: if the method only exploited natural-image shortcuts, its gains would shrink sharply after the source domain changes.

\Cref{tab:patternnet_cross_domain} reports $\text{KMR}_a/\text{KMR}_b/\text{KMR}_c$ and ASR on GPT-4o, Claude 3.7, and Gemini 2.5. Across both datasets, \ours{} consistently surpasses \attackname{} and prior baselines. On \textit{PatternNet}, \ours{} improves Claude 3.7 ASR from 0.48 to 0.73 (+0.25) and raises GPT-4o $\text{KMR}_{a/b/c}$ to 0.83/0.71/0.24. On \textit{ChestMNIST}, the gain is larger on Claude 3.7 (ASR 0.31 $\rightarrow$ 0.83, +0.52), and \ours{} also achieves the highest Gemini 2.5 $\text{KMR}_{a/b/c}$ (0.89/0.76/0.33). The larger gains on Claude under domain shift suggest that stabilizing local-view alignment is especially useful when the target model has weaker prior alignment with the source domain; visually simple medical and overhead images provide fewer natural-image priors, making crop-consistent target semantics more important.

\subsection{Input-Preprocessing Defenses} \label{sec:preproc}
\begin{table*}[tp]
\centering
\renewcommand{\arraystretch}{1.05}
\begin{adjustbox}{width=0.86\textwidth,center}
\footnotesize
\setlength{\tabcolsep}{1.6pt}
\begin{tabular}{c|c|cccc|cccc|cccc}
\toprule
\textbf{Setting} & \textbf{Method} &
\multicolumn{4}{c|}{\textbf{GPT-4o}} &
\multicolumn{4}{c|}{\textbf{Claude 3.7}} &
\multicolumn{4}{c}{\textbf{Gemini 2.5}} \\
\cmidrule(lr){3-6}\cmidrule(lr){7-10}\cmidrule(lr){11-14}
 &  & $\text{KMR}_a$ & $\text{KMR}_b$ & $\text{KMR}_c$ & ASR & $\text{KMR}_a$ & $\text{KMR}_b$ & $\text{KMR}_c$ & ASR & $\text{KMR}_a$ & $\text{KMR}_b$ & $\text{KMR}_c$ & ASR \\
\midrule
\multirow{5}{*}{\textbf{JPEG} ($Q{=}75$)}
 & AttackVLM   & 0.06 & 0.02 & 0.00 & 0.03 & 0.07 & 0.02 & 0.00 & 0.02 & 0.08 & 0.04 & 0.00 & 0.04 \\
 & SSA-CWA    & 0.08 & 0.04 & 0.01 & 0.10 & 0.07 & 0.02 & 0.00 & 0.05 & 0.09 & 0.06 & 0.01 & 0.09 \\
 & AnyAttack   & 0.06 & 0.03 & 0.00 & 0.05 & 0.04 & 0.01 & 0.00 & 0.03 & 0.08 & 0.03 & 0.00 & 0.05 \\
 & \attackname{} & 0.76 & 0.54 & 0.16 & 0.91 & 0.28 & 0.17 & 0.03 & 0.34 & \textbf{0.75} & 0.51 & 0.11 & 0.76 \\
 & \cellcolor[gray]{0.9}\ours{} &
   \cellcolor[gray]{0.9}\textbf{0.89} & \cellcolor[gray]{0.9}\textbf{0.69} & \cellcolor[gray]{0.9}\textbf{0.20} & \cellcolor[gray]{0.9}\textbf{0.97} &
   \cellcolor[gray]{0.9}\textbf{0.55} & \cellcolor[gray]{0.9}\textbf{0.36} & \cellcolor[gray]{0.9}\textbf{0.09} & \cellcolor[gray]{0.9}\textbf{0.68} &
   \cellcolor[gray]{0.9}\textbf{0.75} & \cellcolor[gray]{0.9}\textbf{0.56} & \cellcolor[gray]{0.9}\textbf{0.18} & \cellcolor[gray]{0.9}\textbf{0.82} \\
\midrule
\multirow{5}{*}{\textbf{DiffPure} ($t{=}25$)}
 & AttackVLM   & 0.05 & 0.02 & 0.00 & 0.01 & 0.05 & 0.02 & 0.00 & 0.01 & 0.08 & 0.03 & 0.00 & 0.01 \\
 & SSA-CWA     & 0.07 & 0.03 & 0.00 & 0.02 & 0.04 & 0.02 & 0.00 & 0.03 & 0.07 & 0.01 & 0.00 & 0.05 \\
 & AnyAttack   & 0.07 & 0.03 & 0.00 & 0.04 & 0.02 & 0.02 & 0.00 & 0.04 & 0.09 & 0.04 & 0.00 & 0.07 \\
 & \attackname{} & 0.42 & 0.20 & 0.03 & 0.43 & 0.10 & 0.05 & 0.01 & 0.10 & 0.39 & 0.22 & 0.01 & 0.32 \\
 & \cellcolor[gray]{0.9}\ours{} &
   \cellcolor[gray]{0.9}\textbf{0.73} & \cellcolor[gray]{0.9}\textbf{0.47} & \cellcolor[gray]{0.9}\textbf{0.15} & \cellcolor[gray]{0.9}\textbf{0.72} &
   \cellcolor[gray]{0.9}\textbf{0.19} & \cellcolor[gray]{0.9}\textbf{0.11} & \cellcolor[gray]{0.9}\textbf{0.04} & \cellcolor[gray]{0.9}\textbf{0.20} &
   \cellcolor[gray]{0.9}\textbf{0.61} & \cellcolor[gray]{0.9}\textbf{0.42} & \cellcolor[gray]{0.9}\textbf{0.06} & \cellcolor[gray]{0.9}\textbf{0.56} \\
\midrule
\multirow{5}{*}{\textbf{DiffPure} ($t{=}75$)}
 & AttackVLM   & 0.08 & 0.05 & 0.00 & 0.02 & 0.04 & 0.02 & 0.00 & 0.00 & 0.04 & 0.01 & 0.00 & 0.01 \\
 & SSA-CWA     & 0.05 & 0.03 & \textbf{0.01} & 0.06 & 0.05 & \textbf{0.03} & 0.00 & 0.03 & 0.07 & 0.02 & 0.00 & 0.05 \\
 & AnyAttack   & 0.05 & 0.00 & 0.00 & 0.06 & 0.04 & 0.02 & 0.00 & 0.03 & 0.04 & 0.02 & 0.00 & 0.07 \\
 & \attackname{} & 0.10 & 0.02 & 0.00 & 0.04 & 0.03 & 0.02 & 0.00 & 0.02 & 0.05 & 0.05 & 0.00 & 0.05 \\
 & \cellcolor[gray]{0.9}\ours{} &
   \cellcolor[gray]{0.9}\textbf{0.13} & \cellcolor[gray]{0.9}\textbf{0.06} & \cellcolor[gray]{0.9}\textbf{0.01} & \cellcolor[gray]{0.9}\textbf{0.07} &
   \cellcolor[gray]{0.9}\textbf{0.07} & \cellcolor[gray]{0.9}0.02 & \cellcolor[gray]{0.9}0.00 & \cellcolor[gray]{0.9}\textbf{0.06} &
   \cellcolor[gray]{0.9}\textbf{0.12} & \cellcolor[gray]{0.9}\textbf{0.06} & \cellcolor[gray]{0.9}\textbf{0.01} & \cellcolor[gray]{0.9}\textbf{0.08} \\
\bottomrule
\end{tabular}
\end{adjustbox}
\caption{Robustness under input-preprocessing defenses. We report KMR$_{a/b/c}$ and ASR (higher is better) for GPT-4o, Claude 3.7, and Gemini 2.5. Bold indicates the best value within each metric column for the given defense block; shaded cells highlight \ours{}.}
\label{tab:preproc_unified}
\end{table*}

We evaluate two input-preprocessing defenses: JPEG recompression (quality $Q{=}75$) and diffusion-based purification (DiffPure) with denoising budgets $t{=}25$ and $t{=}75$. As summarized in \cref{tab:preproc_unified}, \ours{} remains strong under JPEG while prior attacks substantially degrade, suggesting resilience to quantization and mild photometric shifts. DiffPure reduces success rates for all methods; however, \ours{} preserves a clear margin at $t{=}25$ and remains the most effective even under the aggressive $t{=}75$ setting, where purification approaches image regeneration.

\subsection{Human Perceptual Study}\label{subsec:human_perceptual}
\begin{table*}[tp]
\centering
\renewcommand\arraystretch{1.12}
\small
\begin{minipage}[b]{0.48\textwidth}\centering
\setlength{\tabcolsep}{4pt}
\begin{tabular}{lcc}
\toprule
Proportion & $\epsilon=16$ & $\epsilon=8$ \\
\midrule
Adversarial images identified & $42 \pm 1.7$ & $27.4 \pm 1.6$ \\
Original images identified & $98 \pm 1.6$ & $93.1 \pm 2.3$ \\
\bottomrule
\end{tabular}
\end{minipage}\hfill
\begin{minipage}[b]{0.48\textwidth}\centering
\setlength{\tabcolsep}{6pt}
\begin{tabular}{lc}
\toprule
Method & Perturbed (\%) \\
\midrule
AnyAttack   & $84 \pm 4.47$ \\
SSA-CWA     & $54 \pm 8.49$ \\
\attackname{} & $30 \pm 10.1$ \\
\ours{} & $32 \pm 8.0$ \\
\bottomrule
\end{tabular}
\end{minipage}

\vspace{0.4em}
\begin{minipage}[t]{0.48\textwidth}
\caption{Human study on the imperceptibility of \ours{} under different perturbation budgets. We report the proportion (\%) of images identified by users; results are averaged over 10 user groups (mean~$\pm$~std).}
\label{tab:user-study-eps}
\end{minipage}\hfill
\begin{minipage}[t]{0.48\textwidth}
\caption{Proportion of adversarial images from each attack that participants judged as perturbed (all at $\epsilon=16$). Lower values indicate more perceptually stealthy perturbations.}
\label{tab:user-study-methods}
\end{minipage}
\end{table*}

We conducted two user studies to evaluate the perceptual stealth of the perturbations beyond $\ell_p$ norms.

\paragraph{\ours{} against clean images.}
Participants were shown 50 images (25 perturbed by \ours{} and 25 unmodified) in a random order and asked to label each image as ``perturbed'' or ``clean''. Adversarial images were generated with a perturbation budget $\epsilon = 16$ or $\epsilon = 8$. \Cref{tab:user-study-eps} summarizes the results, averaged over 10 distinct user groups. At $\epsilon=16$, only 42\% of \ours{} adversarial images are identified as perturbed, so 58\% pass the human check even when participants are told to look for perturbations. At $\epsilon=8$, the identified proportion drops to 27.4\%, and the accuracy on clean images also decreases.

\paragraph{Comparison across attack methods.}
Each of 10 participants was shown 40 images: 10 adversarial examples from each of AnyAttack, SSA-CWA, \attackname{}, and \ours{}, all at $\epsilon=16$. Participants were told that exactly half of the images were perturbed and asked to select the 20 images they believed were most perturbed. As reported in \cref{tab:user-study-methods}, AnyAttack is the most easily detected method (84\%), SSA-CWA follows (54\%), while \attackname{} and \ours{} are flagged only about 30\% of the time. The slight difference in $\ell_p$ norms between \attackname{} and \ours{} does not translate into a difference in human perceptibility.

\subsection{Runtime and Cost Analysis}
\label{subsec:cost}

Each iteration of \ours{} back-propagates through the $K$ source crops and only forward-propagates the $P$ auxiliary targets. Since a backward pass is roughly twice as expensive as a forward pass, the per-iteration complexity is \(\mathcal{O}\!\bigl(K\,(3+P)\bigr)\), with a raw per-crop cost factor of $(3+P)/3$ over source-only back-propagation. More concretely, let $d$ denote the hidden dimension and $N$ the sequence length. One Transformer layer has approximate forward cost $M \triangleq 12Nd^{2}+4N^{2}d$, and one forward--backward update costs about $3M$. With $\rho_N$ denoting the measured per-iteration FLOP ratio of an $N$-model ensemble to one CLIP-B/16 model, AttackVLM costs $3M$, \attackname{} costs $3\rho_NM$, SSA-CWA costs $3\rho_N\hat K M$ with $\hat K$ inner sharpness-aware steps, and \ours{} costs $\rho_NK(3+P)M$.

Wall-clock cost is milder because crops and auxiliary forwards are batched. On a single NVIDIA RTX~4090 GPU with batch size 32, SSA-CWA costs $545.80\pm4.21$\,s per image, \attackname{} costs $22.04\pm0.11$\,s, and \ours{} with $K=2,P=2$ costs $24.13\pm0.84$\,s. This is a $9.4\%$ increase over \attackname{} while improving Claude~3.7 by $+20\%/+17\%/+3\%/+13\%$ on KMR$_a$/KMR$_b$/KMR$_c$/ASR and GPT-4o by $+8\%/+3\%/+1\%/+6\%$. The cost profile is dominated by batched surrogate evaluation, so the wall-clock increase is much smaller than the raw crop count may suggest. We omit AttackVLM and AnyAttack from the runtime comparison: AttackVLM reports low attack performance, and AnyAttack spends most of its compute in a separate pre-training stage.

\section{Qualitative Results} \label{appendix:vis}
\begin{figure*}[tp]
    \centering
    \includegraphics[width=0.86\textwidth]{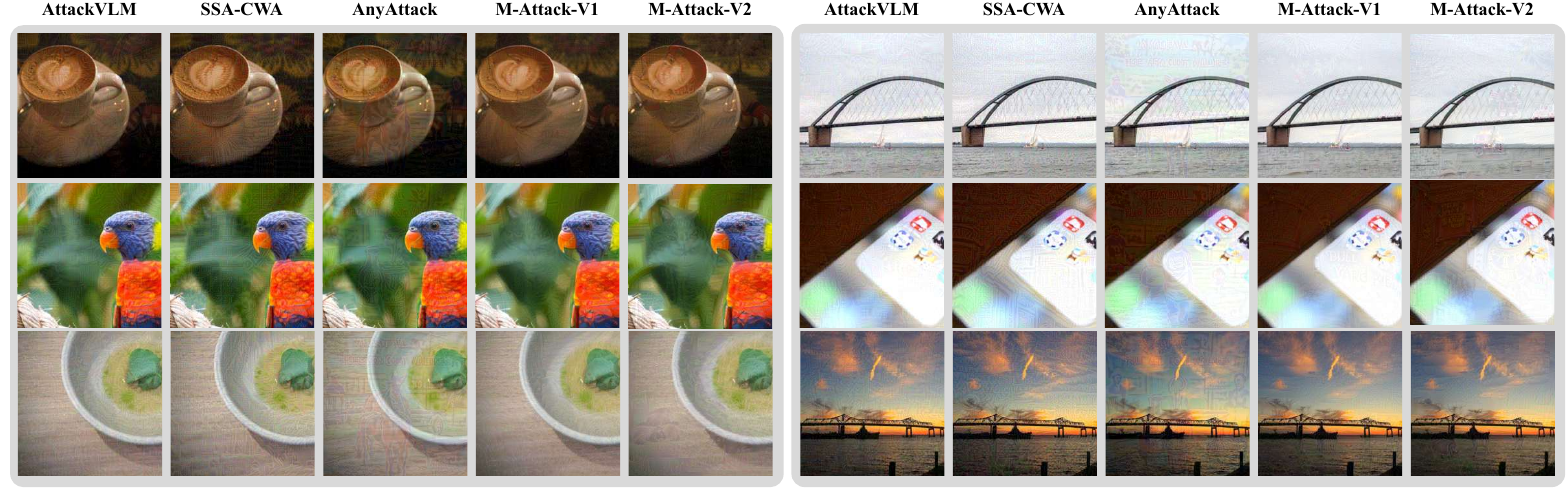}
    \caption{Visualization of adversarial samples under $\epsilon=8$.}
    \label{fig:epsilon 8 full}
\end{figure*}
\begin{figure*}[tp]
    \centering
    \includegraphics[width=0.86\textwidth]{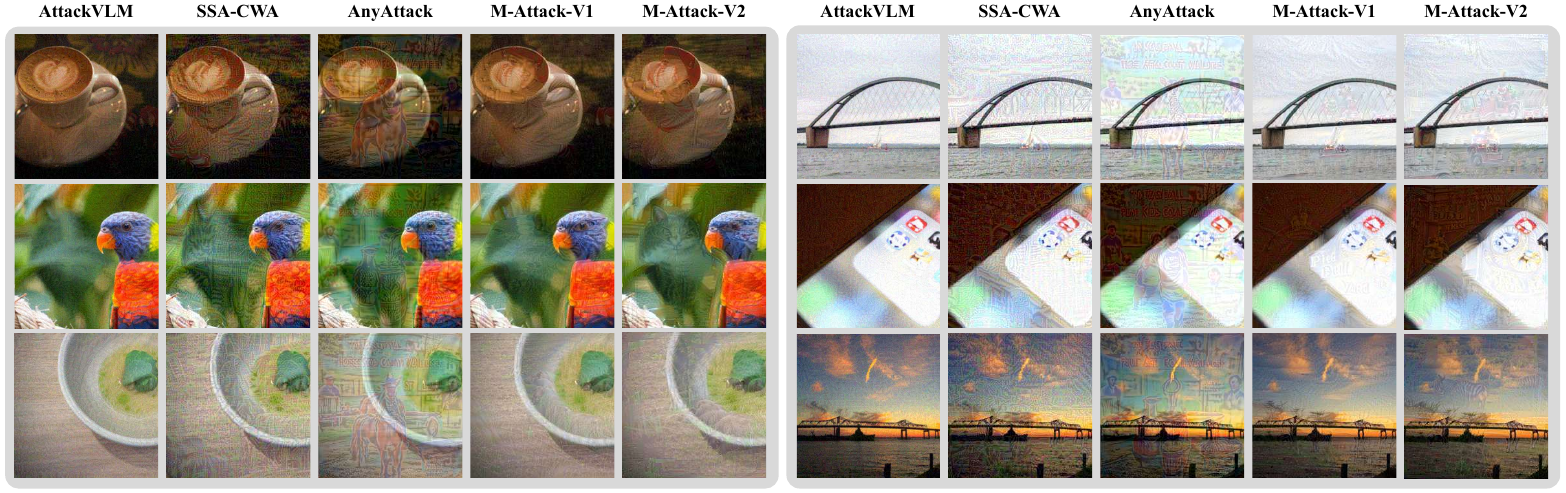}
    \caption{Visualization of adversarial samples under $\epsilon=16$.}
    \label{fig:epsilon 16}
\end{figure*}
\begin{figure*}[tp]
    \centering
    \includegraphics[width=0.95\textwidth]{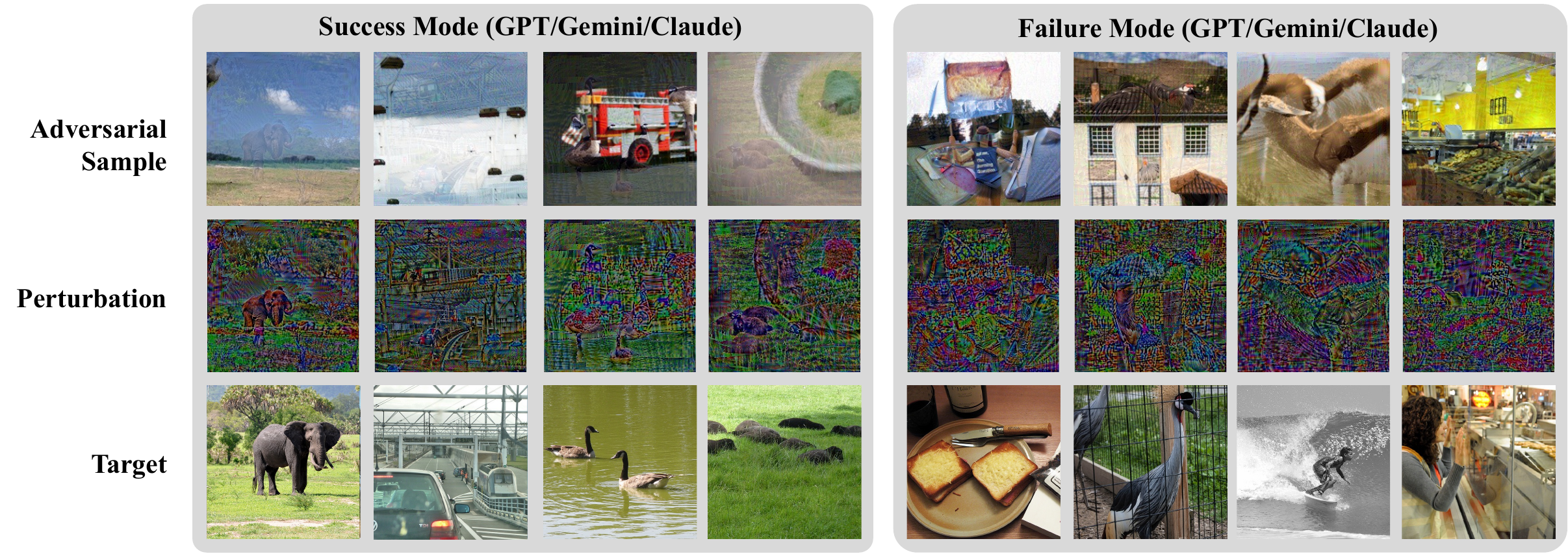}
    \caption{Success and failure modes of \ours{} shared by GPT-5, Claude 4.0-extended, and Gemini 2.5-Pro.}
    \label{fig:success_failure}
\end{figure*}
\begin{figure*}[tp]
    \centering
    \includegraphics[width=0.7\textwidth]{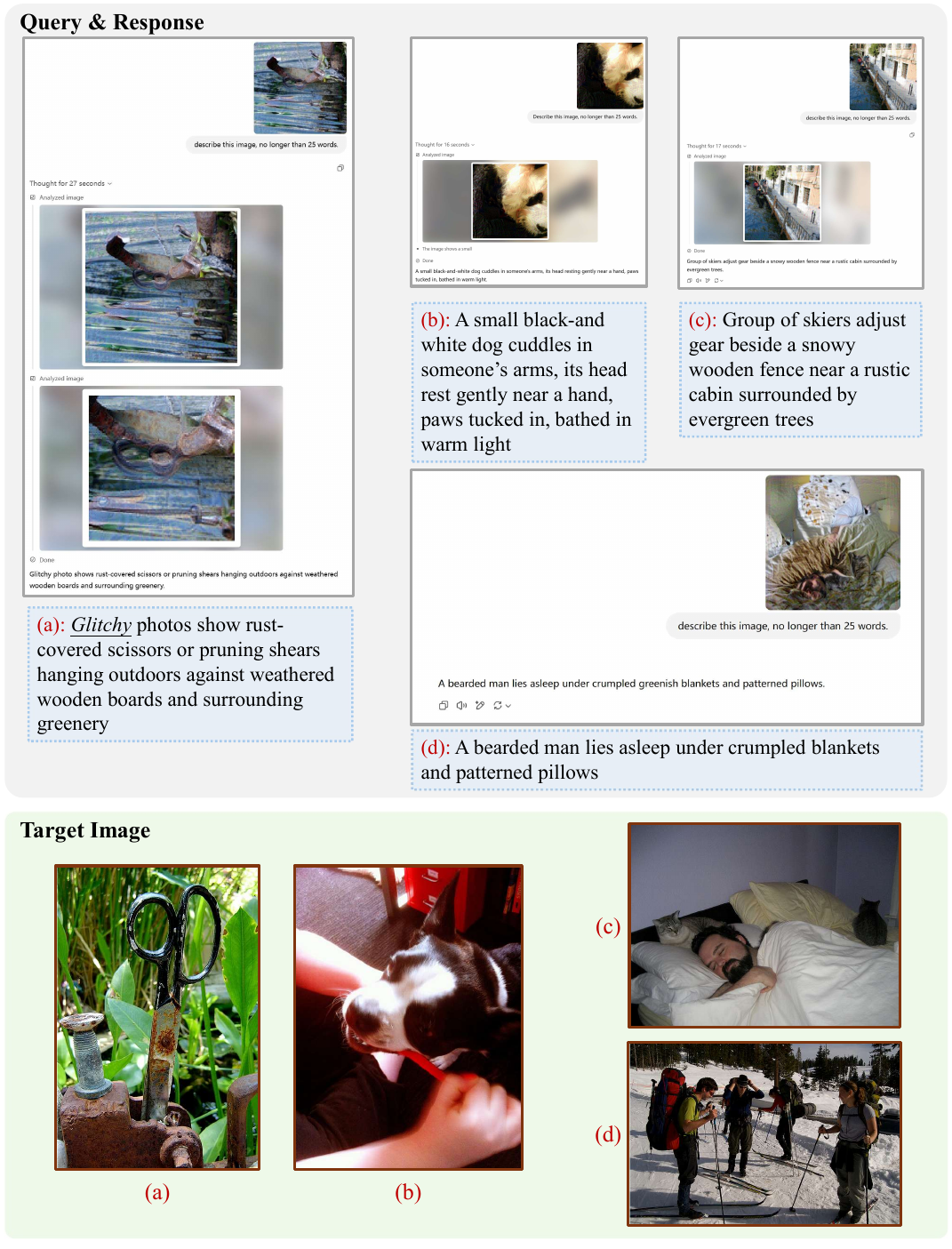}
    \caption{Visualization of GPT-o3's responses to \ours{} adversarial samples. The underlined `glitchy' denotes that GPT-o3 notices something unusual.}
    \label{fig:response}
\end{figure*}

\paragraph{Adversarial examples.} \Cref{fig:epsilon 8 full,fig:epsilon 16} visualize adversarial samples of different black-box attacks under $\epsilon=8$ and $\epsilon=16$. The visual effect of \ours{} is very close to that of \attackname{} under both budgets.

\paragraph{Success and failure modes.} \Cref{fig:success_failure} illustrates typical successful and failed attacks of \ours{} across GPT, Claude, and Gemini. Successful cases show clearer target-related shapes in the perturbation and tend to use neater, more centralized targets, suggesting that visually coherent references provide more stable alignment anchors.

\paragraph{Vision-reasoning responses.} \Cref{fig:response} shows how GPT-o3 responds to \ours{} adversarial samples. Its behavior falls into three types: no reasoning, simple reasoning, and zoom-in reasoning. In the zoom-in case, GPT-o3 flags the central area as uncertain and inspects it, yet its final answer remains semantically close to the target image.

\end{document}